%% file: segmentation.tex
\documentclass[journal]{IEEEtran}
\input{commands}

\usepackage{xcolor}
\usepackage{colortbl}
\usepackage{algorithm}
\usepackage{algorithmic}
\usepackage{bbm}
\usepackage{mathtools}
\newcommand{\etal}{{\it{et al}.} }
\DeclarePairedDelimiter\floor{\lfloor}{\rfloor}

\begin{document}
\title{Depth Adaptive Deep Neural Network \\for Semantic Segmentation}

\author{Byeongkeun~Kang,
        Yeejin~Lee,
        and~Truong~Q.~Nguyen,~\IEEEmembership{Fellow,~IEEE}
\thanks{B. Kang, Y. Lee, and T. Q. Nguyen are with the Department of Electrical and Computer Engineering, University of California, San Diego, CA 92093 USA (e-mail: bkkang@ucsd.edu, yel031@ucsd.edu, tqn001@eng.ucsd.edu).}
\thanks{This work is supported in part by NSF grant IIS-1522125.}
}
\maketitle

\begin{abstract}
In this work, we present the depth-adaptive deep neural network using a depth map for semantic segmentation. Typical deep neural networks receive inputs at the predetermined locations regardless of the distance from the camera. This fixed receptive field presents a challenge to generalize the features of objects at various distances in neural networks. Specifically, the predetermined receptive fields are too small at a short distance, and vice versa. To overcome this challenge, we develop a neural network which is able to adapt the receptive field not only for each layer but also for each neuron at the spatial location. To adjust the receptive field, we propose the depth-adaptive multiscale (DaM) convolution layer consisting of the adaptive perception neuron and the in-layer multiscale neuron. The adaptive perception neuron is to adjust the receptive field at each spatial location using the corresponding depth information. The in-layer multiscale neuron is to apply the different size of the receptive field at each feature space to learn features at multiple scales. The proposed DaM convolution is applied to two fully convolutional neural networks. We demonstrate the effectiveness of the proposed neural networks on the publicly available RGB-D dataset for semantic segmentation and the novel hand segmentation dataset for hand-object interaction. The experimental results show that the proposed method outperforms the state-of-the-art methods without any additional layers or pre/post-processing.
\end{abstract}

\

\IEEEpeerreviewmaketitle

\input{1_introduction}
\input{2_related_works}
\input{3_proposed_method}

\input{3_1_notation}
\input{3_2_adaptive_perception}
\input{3_3_multiscale}

\input{3_4_architecture}
\input{3_5_backpropagating}
\input{3_6_proof}
\input{4_experiments_results}

\section{Conclusion}
In this paper, we presented the novel fully convolutional neural networks that adjust the receptive field using depth information to learn/extract depth-invariant feature representations. In the proposed neural networks, we introduced the DaM convolution layer consisting of the adaptive perception neuron and the in-layer multiscale neuron. The proposed neural networks were applied to indoor semantic segmentation and hand segmentation for hand-object interaction. The experimental results demonstrate that the proposed neural networks improve the accuracy of segmentation without any additional layers or pre/post-processing. 

\section{Acknowledgment}
This work is supported in part by NSF grant IIS-1522125. We thank Nan Jiang for her contribution to data preprocessing and our colleagues for their participation in data collection. We also thank the anonymous reviewers for their insightful comments.

\bibliographystyle{IEEEtran}
\bibliography{segmentation}

\vfill
\end{document}

%% file: commands.tex
\usepackage{subfigure}
\usepackage{cite}
\usepackage[pdftex]{graphicx}
\graphicspath{{../pdf/}{../jpeg/}}
\DeclareGraphicsExtensions{.pdf,.jpeg,.png}
\usepackage{amsmath}
\usepackage{algorithmic}
\usepackage{array}
\ifCLASSOPTIONcompsoc
  \usepackage[caption=false,font=normalsize,labelfont=sf,textfont=sf]{subfig}
\else
  \usepackage[caption=false,font=footnotesize]{subfig}
\fi
\usepackage{url}

\usepackage{amssymb}
\usepackage{multirow}
\usepackage{tabu}
\usepackage{pbox}

{\begin{list}               
    {$\bullet$ \hfill}{
        \setlength{\leftmargin}{\parindent}
        \setlength{\parsep}{0.04\baselineskip}
        \setlength{\itemsep}{0.5\parsep}
        \setlength{\labelwidth}{\leftmargin}
        \setlength{\labelsep}{0em}}
    }
{\end{list}}

\providecommand{\bb}[1]{\textcolor{blue}{\textbf{#1}}}

\providecommand{\cref}[1]{Chapter~\ref{#1}}
\providecommand{\sref}[1]{Section~\ref{#1}}
\providecommand{\fref}[1]{Fig.~\ref{#1}}
\providecommand{\tref}[1]{Table~\ref{#1}}

\providecommand{\R}{\ensuremath{\mathbb{R}}}

\providecommand{\Z}{\ensuremath{\mathbb{Z}}}

\renewcommand{\vec}[1]{\ensuremath{\boldsymbol{#1}}}
\providecommand{\mat}[1]{\ensuremath{\boldsymbol{#1}}}

\providecommand{\calL}{\mathcal{L}}

\providecommand{\calT}{\mathcal{T}}

\providecommand{\mD}{\mat{D}}

\providecommand{\mL}{\mat{L}}

\providecommand{\mS}{\mat{S}}

\providecommand{\mW}{\mat{W}}
\providecommand{\mX}{\mat{X}}

\providecommand{\vb}{\vec{b}}

\providecommand{\vs}{\vec{s}}

\providecommand{\vw}{\vec{w}}
\providecommand{\vx}{\vec{x}}

\providecommand{\vxh}{\hat{\vx}}

\providecommand{\vsh}{\hat{\vs}}



%% file: 1_introduction.tex
\section{Introduction}
\IEEEPARstart{D}{epth} perception, which is one of the crucial abilities in the human visual system, allows human to perceive the distance to an object and to understand the world in three dimensions. The human visual system uses the perceived depth information to robustly estimate the size and shape of objects in three dimensions. The three-dimensional information helps to better understand the objects and scenes along with other cues such as color information. Thus, depth information plays a key role in understanding the visual world.

\begin{figure}[t]
\begin{center}
   \includegraphics[width=0.9\linewidth]{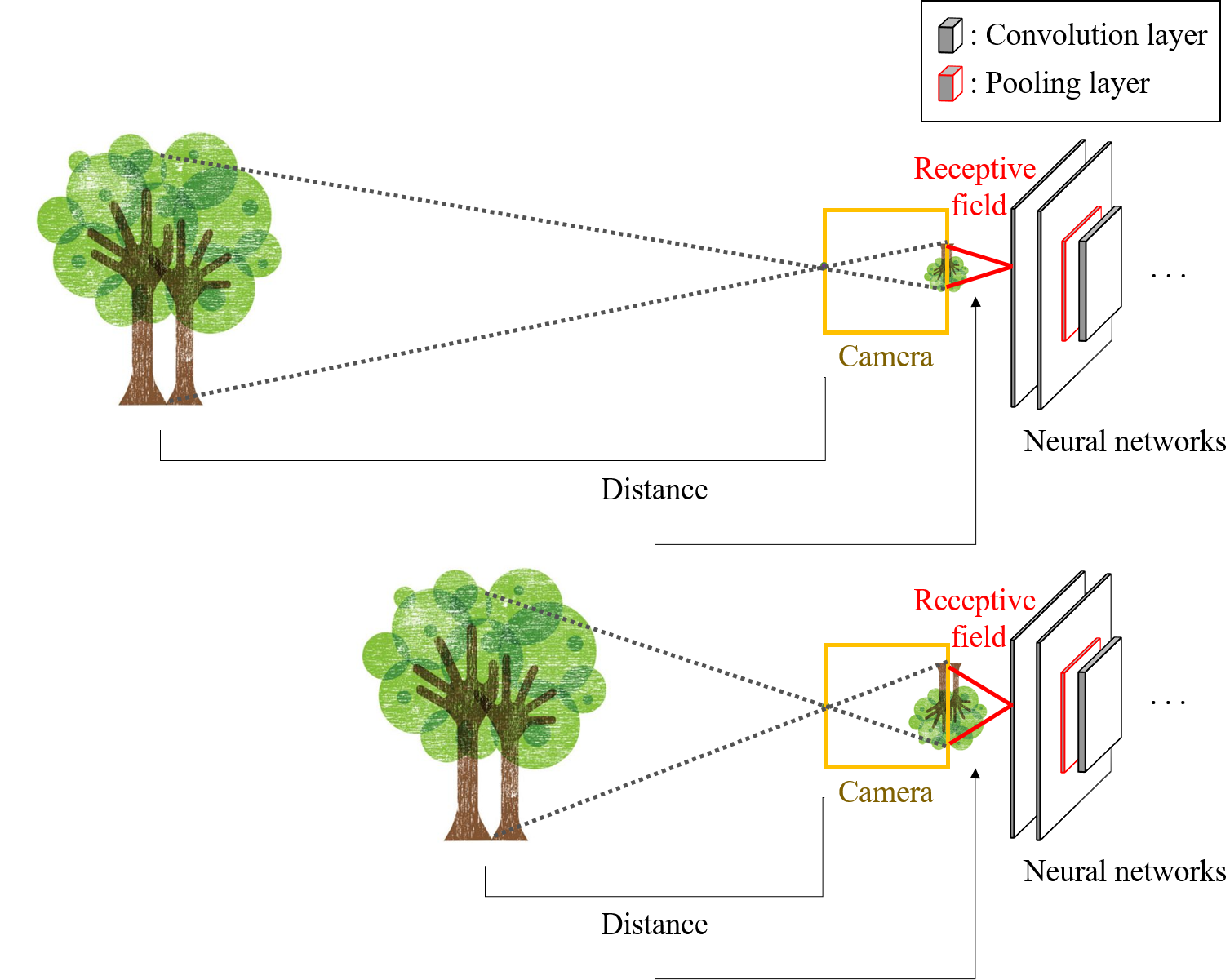}
\end{center}
    \caption{Illustration of the captured images and the proposed neural networks. The size of a captured object on the image plane varies with the distance from the object to the camera.}
\label{fig:pinholeCamera}
\end{figure}

As depth information is crucial for understanding the visual world, many researches have been explored ways to acquire accurate depth information efficiently in both hardware systems and software systems. In hardware-based solutions, advanced depth sensors, such as Microsoft Kinect and light detection and ranging~(LiDAR) sensors have been developed to capture better quality depth information with portability and low cost~\cite{kinect, lidar, lidar2}. In software-based solutions, disparity estimation algorithms using single or multiple cameras have been studied to estimate accurate depth cues in shorter processing time~\cite{disparity1, disparity2, disparity3}. Owing to these successes in both communities, depth information has been widely usable in many computer vision applications such as human pose estimation~\cite{shottoncvpr, shottonpami}, indoor scene understanding~\cite{nyudv2}, and autonomous driving~\cite{kitti, cityscapes}.

After perceiving depth and/or color information, a machine processes the perceived information to understand the visual world. One of the recent popular frameworks for learning visual information is the deep neural network, which is loosely inspired by the neurons of a human brain. As computing capability of machines has increased drastically, deep neural networks have attained a huge improvement in understanding visual information and shown the state-of-the-art performance in many tasks such as image classification~\cite{AlexNet, vgg, resnet}, object detection~\cite{RCNN, fastRCNN, fasterRCNN, yolo, Tripathi}, and semantic segmentation~\cite{longcvpr, longpami, Chen, Chen16, YuKoltun2016}. 

\begin{figure}[t]
\begin{center}
   \includegraphics[width=0.8\linewidth]{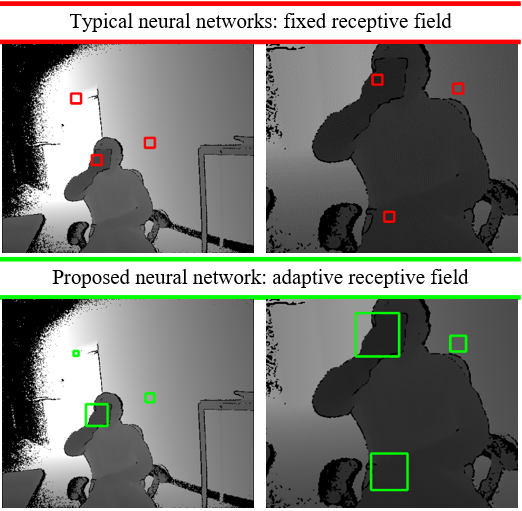}
\end{center}
   \caption{The visual comparison of the receptive fields for typical neural networks and the proposed neural network. The proposed method adjusts the receptive field using the distance from the camera to each pixel. In the proposed neural network, the region of the receptive field is invariant to distances.}
\label{fig:introduction}
\end{figure}

\begin{figure}[t]
\begin{center}
   \includegraphics[width=0.8\linewidth]{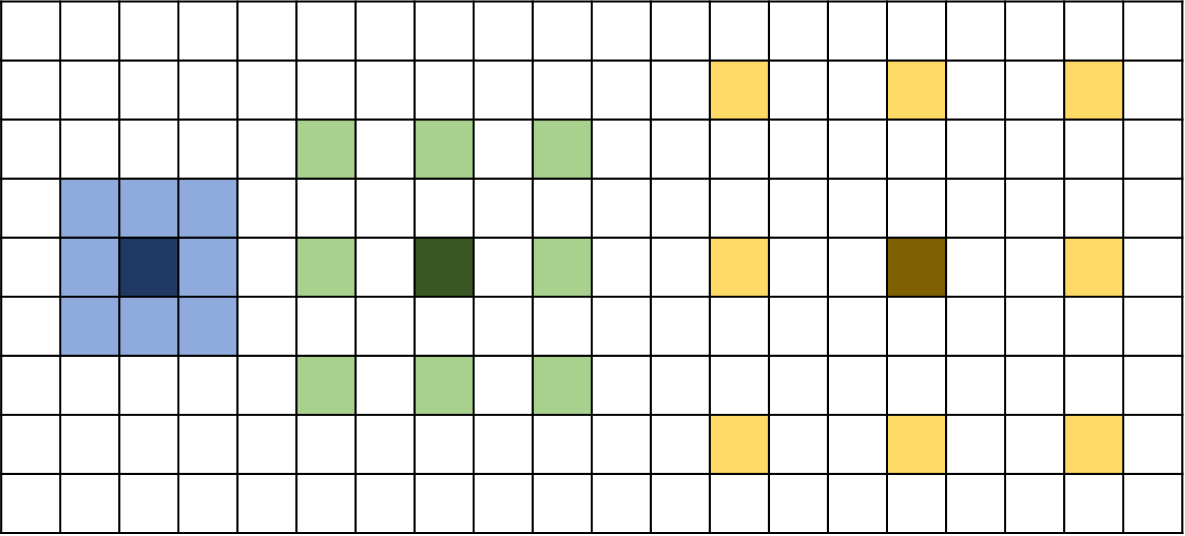}
\end{center}
   \caption{An example of applying the different sizes~(sparsities) $\mS$ of the receptive field at each spatial location $(m,n)$. Suppose the indices of the matrix start from the top-left corner with $(1,1)$, and the kernel size is $3 \times 3$. The figure shows the cases of $\mS^\ell_{5,3}=1$, $\mS^\ell_{5,8}=2$, and $\mS^\ell_{5,16}=3$. }
\label{fig:sparsity}
\end{figure}

Because of the importance of depth information and the improvements by using deep neural networks, it has been speculated that incorporating depth information with neural networks has the advantage in understanding visual information. In most researches on deep neural networks using depth information, the depth map has been treated as an image equivalent input to the networks~\cite{tompson, 3dhandpose, deephand, humanpose, kangacpr}. In such networks, the neurons share the predetermined receptive fields in a convolutional layer, which hinders the networks from learning common representations of an object. Considering that a pinhole camera captures an object at different distances, the camera captures the same object in different sizes, as demonstrated in~\fref{fig:pinholeCamera}. The illustration implies that a neural network can possibly learn/extract different features for the same object at various distances, yielding the confusions of recognizing objects. Hence, we propose the novel deep neural networks that learn common features of the same object by leveraging depth information~(\sref{sec:architecture}). The proposed neural networks perceive the same region of the object regardless of the distance from the camera to each pixel as described in~\fref{fig:introduction}. 
This is achieved by the novel \underline{D}epth-\underline{a}daptive \underline{M}ultiscale or DaM convolution layer consisting of the adaptive perception neuron and the in-layer multiscale neuron in~\sref{sec:adaptiveLayer}. The adaptive perception neuron adjusts the size of the receptive field at each spatial location corresponding to the distance from the camera. The adjustment requires a coefficient to decide the ideal correlation between the size of the receptive field and the distance. Since the optimal coefficient differs depending on the objects, better performance can be achieved by utilizing multiple coefficients in a layer. This is implemented by the proposed in-layer multiscale neuron. The in-layer multiscale neuron learns/extracts diversely scaled representations in a layer by applying a different size of the receptive field at each feature representation. The adjustment of the receptive field is applied using the sparse convolution~(dilated convolution) as demonstrated in~\fref{fig:sparsity}.
In~\sref{sec:result}, we verify the effectiveness of the proposed method on two tasks: indoor semantic segmentation and hand segmentation for hand-object interaction. We use publicly available NYUDv2 dataset~\cite{nyudv2} for indoor semantic segmentation and collect a new challenging dataset including hand-object interaction for hand segmentation.

In summary, the contributions of our work are as follows:
\begin{itemize}
  \item We develop the depth-adaptive neural networks using the DaM convolution. The DaM convolution consists of the adaptive perception neurons and the in-layer multiscale neurons.
  \item We propose the adaptive perception neuron. The neuron learns/extracts depth-adaptive representations. 
  \item We propose the in-layer multiscale neuron. The neuron learns/extracts variously scaled representations in a convolution layer.
  \item We verify the effectiveness of the proposed networks on the task of semantic segmentation.
\end{itemize}

%% file: 2_related_works.tex
\section{Related Works}
\textbf{Deep neural networks using depth map.}
Most researches of deep neural networks using depth maps treated a raw depth map as an image equivalent. For instance, a raw depth map was given as a direct input to the networks in hand pose estimation~\cite{tompson, 3dhandpose, deephand}, human pose estimation~\cite{humanpose}, and fingerspelling recognition~\cite{kangacpr}.

Alternatively, Gupta \etal proposed the geocentric embedding for a depth map to learn better representations in convolutional neural networks~\cite{Gupta2014}. Specifically, the geocentric embedding encodes horizontal disparity, height above ground, and angle with gravity~(HHA) for each pixel. The work showed that using the HHA encoded images, convolutional neural networks can learn better features for object detection and segmentation.

The networks we introduce are distinct from the works in~\cite{tompson, 3dhandpose, deephand, humanpose, kangacpr, Gupta2014}. First, our proposed method utilizes depth information in convolution layers rather than converting a raw depth map into a better representation in a preprocessing stage. In other words, our method does not require any additional preprocessing to manipulate the raw data. Second, our proposed method can take any type of input~(e.g. color image, depth map, etc.) to learn feature representations by giving the corresponding depth information as shown in~\fref{fig:convLayer}.

\textbf{Semantic segmentation.}
\quad Long \etal proposed fully convolutional neural networks~(FCN) for semantic segmentation by converting fully connected layers to convolution layers in the neural networks for image classification~\cite{longcvpr, longpami}. The networks take an input of arbitrary size and produce the correspondingly-sized output.

Additional efforts have been made to improve the performance in~\cite{Zheng, Chen, Chen16, YuKoltun2016}. Zheng \etal proposed the convolutional neural networks that incorporate the strength of conditional random field~(CRF)-based probabilistic graphical modeling. They formulated CRF as recurrent neural networks~(RNN) and attached the RNN after FCN~\cite{Zheng}. Chen \etal improved semantic segmentation using convolution with up-sampled filters, atrous spatial pyramid pooling, and fully connected CRF~\cite{Chen, Chen16}. Yu \etal proposed an additional context module to aggregate multiscale information without losing resolution~\cite{YuKoltun2016}.

Unlike other methods, our approach improves the performance of neural networks using depth information without adding additional layers. In addition, the proposed networks can incorporate any aforementioned additional layers for further improvement.

\textbf{Hand segmentation for hand-object interaction.}
\quad Most algorithms for hand segmentation segment hands using skin color in color images. Oikonomidis \etal and Romero \etal segmented hands by thresholding skin color in the hue-saturation-value~(HSV) color  space~\cite{oikoiccv, romeroicra, romeroivc}. Wang \etal used the learned probabilistic model constructed from the color histogram of the first frame~\cite{wangtog}. The histogram was generated using super-Gaussian mixture model in~\cite{palmer}. Tzionas \etal processed segmentation of hands using the Gaussian mixture model constructed for skin color~\cite{jones, tzionas}.

However, skin color-based hand segmentation is sensitive to skin pigment difference and light condition variation. Similarly, in the segmentation, hands can be confused with other objects in skin color and other body parts~(e.g. arm, face, etc.). To overcome these limitations, we decided to segment hands using only depth maps. For the experiment, we collected a new dataset with pixel-wise annotations because we were not able to find a publicly available dataset for hand-object interaction with annotations.

\textbf{Convolution layer.}
\quad Conventionally, most convolutional neural networks used typical, dense, and fixed convolution~\cite{AlexNet, vgg}. Recently, dilated~(atrous) convolution was applied for semantic segmentation to extract sparse features in higher resolution~\cite{Chen}. The structure excluded pooling layers~(which cause the decrement of spatial resolution) and replaced typical convolutions following the pooling layers by dilated convolutions. The dilated convolution was employed to increase the size of receptive fields and compensate the exclusion of pooling layers~\cite{Chen, Chen16, YuKoltun2016}. The dilated convolution in these methods has different sparsity at each layer depending on the excluded pooling layers while it has the same sparsity for all spatial locations and all feature representations in a layer.

Contemporarily, active convolution and deformable convolution are presented in~\cite{activeConv, deformConv}. The goal of both methods is to learn the shape of convolutions using a training dataset. Active convolution defines the learnable position parameters to represent various forms of the receptive fields in the task of image classification~\cite{activeConv}. The position parameters are shared across all kernels in a layer. Thus, the learned receptive field is the same at all spatial locations and for all feature representations. Deformable convolution uses the offset field similar to the position parameters~\cite{deformConv}. The offset field is computed using the input feature map and has different receptive fields at each spatial location. This deformable convolution was tested on semantic segmentation task and object detection task.

In the proposed networks, we apply dilated~(sparse) convolution~\cite{wavelet} to adjust the size of receptive fields for two purposes. First, we adapt the sparsities in convolutions to learn/extract near depth-invariant representations using distance information. Thus, the sparsity is adjusted at each spatial location depending on the distance at the location. Second, we adapt the sparsity at each feature representation to learn variously scaled representations. That is, the proposed dilated convolution generates different sparsities \emph{at each spatial location} and \emph{at each feature representation}.

%% file: 3_proposed_method.tex
\section{Proposed Method}
The goal of this work is to learn depth-invariant representations in deep neural networks using depth information. To achieve this goal, we propose the novel DaM convolution layer conceiving the adaptive perception neurons and the in-layer multiscale neurons as described in \fref{fig:convLayer}. The adaptive perception neuron is proposed to adjust the receptive field using the depth information at each spatial location. The in-layer multiscale neuron is designed to learn features in different scales at each feature space~(or channel) in a layer.  

In \sref{sec:notation}, we introduce key notations for networks. We provide the detailed explanation of the DaM convolution consisting of the adaptive perception neuron and the in-layer multiscale neuron in~\sref{sec:adaptiveLayer}. The overall architecture of the proposed neural network is developed in~\sref{sec:architecture}. In \sref{sec:backpropagate}, the details of the training procedure are derived for the proposed networks. Finally, we provide the mathematical proof of the depth invariant property of the proposed networks in \sref{sec:proof}.

%% file: 3_1_notation.tex
\input{fig_conv_layer}

\subsection{Notation} \label{sec:notation}
\begin{figure}[t]
\begin{center}
   \includegraphics[width=0.9\linewidth]{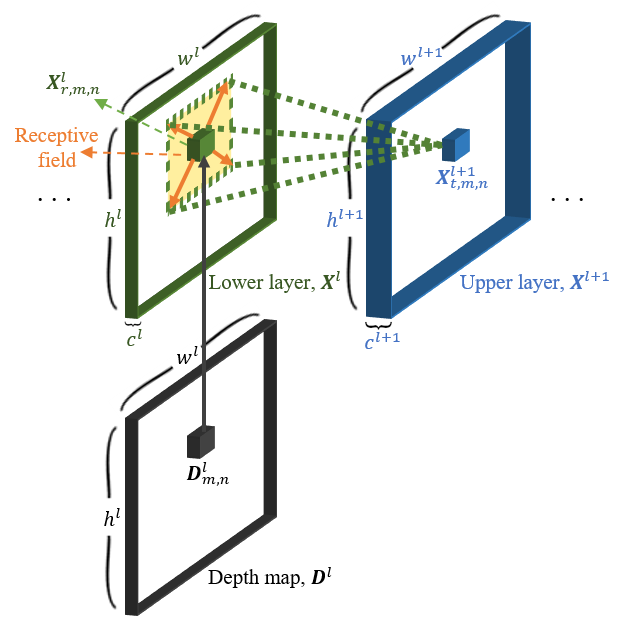}
\end{center}
   \caption{Notations and the adaptive perception neuron. This neuron adjusts the size of the receptive field based on the depth information at each spatial location.}
\label{fig:AdaptiveNeuron}
\end{figure}

Let $\mX^\ell \in \R^{c^\ell \times h^\ell \times w^\ell}$ and $\mX^{\ell+1} \in \R^{c^{\ell+1} \times h^{\ell+1} \times w^{\ell+1}}$ be the matrices representing an input and an output of a certain layer~$\ell$~(either convolution, pooling, softmax, or loss layer), where $c$, $h$, and $w$ denote the number of feature spaces~(channels), height, and width, respectively. Also, let $\mD^\ell \in \R^{ h^\ell \times w^\ell}$ be the (pooled) depth map in the convolution layer~$\ell$ whose spatial resolution corresponds to the spatial resolution of the input $\mX^\ell$~(see Figs.~\ref{fig:convLayer} and \ref{fig:AdaptiveNeuron}). The size of $\mD^\ell$ is determined by pooling, convolution, and padding in the previous layers.

The output $\mX^{\ell+1}$ of the convolution layer $\ell$ is computed by convolving the input $\mX^\ell$ with a shared weight matrix $\mW^\ell \in \R^{c^{\ell+1} \times c^\ell \times k^\ell_h \times k^\ell_w}$ and by adding a bias vector $\vb^\ell \in \R^{c^{\ell+1}}$, where $k^\ell_h$ and $k^\ell_w$ denote the dimensions of kernels along the height and width directions. In a typical convolution layer, the output $\mX^{\ell+1}_{t,m,n}$ of the $t$-th output feature space at the spatial location $(m,n)$ is computed as
\begin{equation}
\mX^{\ell+1}_{t,m,n} = f\bigg(\sum_{r} \sum_{u}\sum_{v}{\mW^\ell_{t,r,u,v} \mX^\ell_{r, m+u,\, n+v}} \:\:+\:\: \vb^\ell_{t}\bigg),
\label{eq:typicalConv}
\end{equation}
where $r \in [0, c^\ell-1]$ and $t \in [0, c^{\ell+1}-1]$ are the indices for the feature spaces of the input and the output, respectively, $u \in [-\floor{k_h/2}, -\floor{k_h/2}+k_h-1]$ and $v \in [-\floor{k_w/2}, -\floor{k_w/2}+k_w-1]$ are the indices for the weight matrix $\mW^\ell$ along the height and width direction, and $f(\cdot)$ is a transfer function~(e.g. rectified linear unit~(ReLu), etc.). 

%% file: fig_conv_layer.tex
\begin{figure}[t]\begin{center}
\begin{minipage}{0.24\linewidth}
\centerline{\includegraphics[scale=0.8]{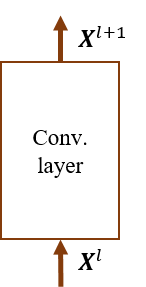}}
\centerline{\footnotesize (a)} 
\end{minipage}
\begin{minipage}{0.74\linewidth}
\centerline{\includegraphics[scale=0.8]{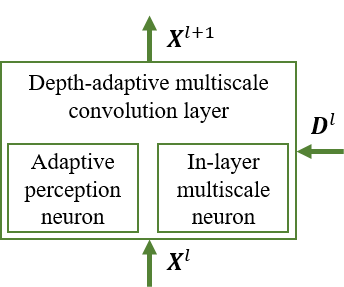}}
\centerline{\footnotesize (b)} 
\end{minipage}
\end{center}
   \caption{(a) Typical convolution layer.  (b) Depth-adaptive multiscale convolution layer.}
\label{fig:convLayer}
\end{figure}

%% file: 3_2_adaptive_perception.tex
\subsection{Depth-adaptive Multiscale Convolution Layer} \label{sec:adaptiveLayer}
As observed in~\fref{fig:pinholeCamera}, an object appears to have different sizes in the image plane depending on its distance from the camera. The generalization performance of the trained networks using these depth-variant features may not be sufficiently good because learning a common representation is challenging from the features. As such, it is necessary to learn depth-invariant features for neural networks in order to achieve better generalization performance. To this end, we propose the DaM convolution layer containing the adaptive perception neurons and the in-layer multiscale neurons. The adaptive perception neuron in~\sref{sec:adaptivePerception} adjusts its receptive field to offset the change of the spatial size of objects on captured images. The receptive field adjusted by the adaptive perception neuron is clearly sub-optimal because the ideal correlation between the size and the distance varies over objects~(e.g. due to different sizes). Hence, we develop the in-layer multiscale neuron in~\sref{sec:multiscale} that effectively controls the size of receptive fields over individual objects. The in-layer multiscale neuron extracts the diversely scaled depth-invariant features by tuning a parameter that determines sparsity at each feature representation.

Given a depth map as an input of the networks, unlike color images, the intensity~(value) of an object on the depth map is scaled by the distance from the camera. This implies that the networks may learn intensity-variant features for the same object. To avoid this misguiding, we propose to employ depth difference~(relative depth) as an input for the feature extraction in~\sref{sec:relativeDepth}.

\subsubsection{Adaptive perception neuron} \label{sec:adaptivePerception}
The proposed adaptive perception neuron determines its size of receptive field based on the depth information at each spatial location while other methods~\cite{Chen, Chen16, YuKoltun2016} used the predetermined receptive field in a convolution layer. Thus, the proposed networks having such adaptive perception neurons can apply different receptive field at each spatial location. Specifically, we increase the receptive field for objects at a close distance and decrease it for objects at a long distance to compensate for the variation of objects' size on the captured images.

To determine the receptive field of each neuron, the depth map $\mD^\ell$ is fed to the adaptive perception neuron. The size of the receptive field $\mS^\ell \in \R^{c^{\scriptscriptstyle \ell} \times h^{\scriptscriptstyle \ell} \times w^{\scriptscriptstyle \ell}}$ at a spatial location $(m,n)$ inversely increases to the depth from the camera $\mD^\ell_{m,n}$, as follows: 
\begin{equation}
\mS^\ell_{r,m,n} \propto \frac{1}{\mD^\ell_{m,n}}.
\label{eq:scale}
\end{equation}

Applying the $\mS^\ell$ for the convolution layer $\ell$, the adaptive perception neuron takes different entries of the input $\mX^\ell$ connected by $\mS^\ell_{r,m,n}$ as demonstrated in Figs.~\ref{fig:sparsity} and \ref{fig:AdaptiveNeuron}. Thus, the output in  (\ref{eq:typicalConv}) is replaced by: 
\begin{equation}
\begin{split}
&\mX^{\ell+1}_{t,m,n} \\
&= f\bigg(\sum_{r} \sum_{u}\sum_{v}{\mW^\ell_{t,r,u,v} \mX^\ell_{r, m+ \mS^\ell_{r,m,n} u,\, n + \mS^\ell_{r,m,n} v}} + \vb^\ell_{t}\bigg).
\label{eq:output1}
\end{split}
\end{equation}

%% file: 3_3_multiscale.tex
\begin{figure}[t]
\begin{center}
   \includegraphics[width=0.9\linewidth]{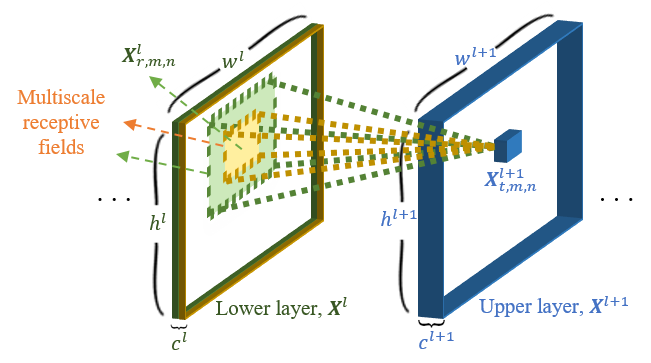}
\end{center}
   \caption{The in-layer multiscale neuron. This neuron is able to learn features at different scales in a layer. }
\label{fig:multiscale}
\end{figure}

\subsubsection{In-layer multiscale neuron} \label{sec:multiscale} 
Conventionally, learning/extracting features in various scales is advantageous in achieving higher segmentation accuracy by learning variant features. To learn features in multiple scales, the neural networks comprised of multiple neural networks were proposed in~\cite{tompson}, known as the multiscale neural networks. In this type of neural networks, each constituting neural network takes an input in different resolution and learns features in various scales. However, these networks are structurally complex and require higher computational complexity.
Thus, we propose the in-layer multiscale neuron that takes only an input and learns features with multiple scales in a network~(see \fref{fig:multiscale}). The proposed in-layer multiscale neuron learns features at various scales by having a different parameter for the sparsity at each feature representation~(channel). 
 
The in-layer multiscale neuron determines sparsity at each feature space $r$ using the multiscale parameter $p^{\ell}_r$, whereas the adaptive perception neuron in the previous section spatially determines the sparsity depending on the depth $\mD^{\ell}_{m,n}$. The parameter $p^{\ell}_r$ is determined as follows:
\begin{equation}
p^{\ell}_r = \frac{s_r^\ell}{\prod_{\ell' \in \calL} z^{\ell'}} \cdot \left[ \frac{1}{|\calT| h^\ell w^\ell} \sum_{d \in \calT} \sum_{m} \sum_{n} {\mD_{m,n}^d} \right]  \cdot q^\ell 
\label{eq:multiscale}
\end{equation}
where $s_r^\ell$ is the scaling factor for each feature space~(channel) $r$, $z^{\ell'}$ is the stride of pooling layers $\ell' \in \calL$ up to the current layer, $|\calT|$ represents the number of data in the training dataset $\calT$, and $q^\ell$ is the dilation parameter from the ancestor architecture.

The $p^{\ell}_r$ is interpreted as three factors: one is the scaling factor $s_r^\ell$ with the mean $\left[ \frac{1}{|\calT| h^\ell w^\ell} \sum_{d \in \calT} \sum_{m} \sum_{n} {\mD_{m,n}^d} \right]$ of the depth maps in the training dataset, another is the factor $1/z^{\ell'}$ regarding pooling layers, and the other is the dilation parameter $q^\ell$ from the ancestor architecture. The scaling factor $s_r^\ell$ with the mean of the depth determines different sparsities at each feature space considering the mean of the depth. Precise parameters for $s_r^\ell$ is explained in~\sref{sec:result}. The term $1/z^{\ell'}$ compensates for the decrement of the spatial resolution of the feature map, caused by pooling layers. That is, the size of the receptive field is decreased as pooling layer reduces the spatial resolution. The term $q^\ell$ is to retain the dilation parameter from the ancestor architecture.

Finally, the size of receptive field is determined by incorporating adaptive perception neuron and in-layer multiscale neuron. The size $\mS^\ell_{r,m,n}$ at a feature space $r$ and a spatial location $(m,n)$ is as follows:
\begin{equation}
\mS^\ell_{r,m,n} = \frac{p^{\ell}_r}{\mD^\ell_{m,n}},
\label{eq:scale}
\end{equation}
where denominator is contributed by the adaptive perception neuron, and numerator is from the in-layer multiscale neuron.

\subsubsection{Depth difference} \label{sec:relativeDepth}
In practice, values on a depth map vary as the distance from the camera changes. For instance, objects at different distances are represented by different intensity levels. However, the relative distance between these objects is constant regardless of their distance from the camera~\cite{shottoncvpr, shottonpami, kangGlobalSIP}. Consequently, we instead use the relative depth to measure distance-independent depth in the first convolution layer.
The relative depth is computed as the difference between the depth at the receptive field and the depth at the center location of the receptive field. Replacing a depth by the relative depth, \eqref{eq:output1} is rewritten as
\begin{equation}
\begin{split}
\label{eq:relativeDepth}
\mX^{2}_{t,m,n} =& f\bigg(\sum_{r} \sum_{u} \sum_{v} \mW^1_{t,r,u,v} \\
&(\mX^1_{r, m+\mS^1_{r,m,n}u,\, n+\mS^1_{r,m,n}v} - \mX^1_{r,m,n}) + \vb^1_{t}\bigg).
\end{split}
\end{equation}
Although the input $\mX^1$ to the networks is replaced by the relative depth $(\mX^1_{r, m+\mS^1_{r,m,n}u,\, n+\mS^1_{r,m,n}v} - \mX^1_{r,m,n})$, the size $\mS^1$ of the receptive fields is computed using the raw depth map $\mD^1$.

%% file: 3_4_architecture.tex
\subsection{Architecture} \label{sec:architecture}
The proposed DaM convolution layer is applied to all convolution layers in two fully convolutional neural networks~(Frontend module~\cite{YuKoltun2016} and DeepLab~\cite{Chen16}). All original convolution layers are replaced by the proposed layers to achieve depth-invariance as demonstrated in \sref{sec:proof}. Frontend module and DeepLab are selected as our baseline model since they are two of the state-of-the-art methods. For DeepLab, we employed the VGG-16~\cite{vgg} network-based architecture with large atrous spatial pyramid pooling~(ASPP-L) and without conditional random field~(CRF)~\cite{Chen16}.

We train the proposed neural networks by back-propagating the multinomial logistic loss $e_{\scriptscriptstyle a}$ while penalizing the increment of weights using the $L_2$ regularization~(denote as $e_{\scriptscriptstyle b}$)~\cite{PRML}. Thus, the total loss $e$ is the weighted sum of $e_{\scriptscriptstyle a}$ and $e_{\scriptscriptstyle b}$~(i.e. $e = e_{\scriptscriptstyle a}+\lambda e_{\scriptscriptstyle b}$), where $\lambda$ is the decay factor. To compute the multinomial logistic loss $e_{\scriptscriptstyle a}$, we apply the softmax function that transfers the input $\mX^\ell \in \R^{c_t \times h^{\scriptscriptstyle \ell} \times w^{\scriptscriptstyle \ell}}$ from the last convolution layer to the output $\mX^{\ell+1} \in \R^{c_t \times h^{\ell} \times w^{\ell}}$, where $c_t$ denotes the total number of classes. In softmax layer, the spatial resolutions of the input $\mX^\ell$ and the output $\mX^{\ell+1}$ are equivalent~(i.e. $(h^{\ell+1}, w^{\ell+1}) = (h^{\ell}, w^{\ell})$). The softmax output of the $r$-th feature space at the spatial location $(m,n)$ is defined as
\begin{equation}
\mX^{\ell+1}_{r,m,n} = \frac{exp(\mX^\ell_{r,m,n})}{\sum_{r} exp(\mX^\ell_{r,m,n})}.
\label{eq:softmax}
\end{equation}
The output $\mX^{\ell+1}_{r,m,n}$ is equivalent to the predicted probability of being the class $r$ at the spatial location $(m,n)$. Then, the multinomial logistic loss $e_{\scriptscriptstyle a}$ is the weighted sum over the logistic outputs of $\mX^{\ell+1}$:
\begin{equation}
e_a = -\frac{1}{h^{\ell} w^{\ell}} \sum_{r} \sum_{m} \sum_{n} \mathbbm{1}(r=\mL_{m,n}) log(\mX^{\ell+1}_{r,m,n}),
\label{eq:loss}
\end{equation}
where $\mathbbm{1}( \cdot )$ is an indicator function and $\mL \in \mathbb{R}^{h^{\ell+1} \times w^{\ell+1}}$ is a target class label matrix. 

%% file: 3_5_backpropagating.tex
\subsection{Back Propagation}
\label{sec:backpropagate}
To train the proposed networks, the loss $e$ is propagated backward and used to update the weights. The weights are updated by minimizing $e$ using the gradient ${\partial e}/{\partial \mW^\ell}$, where the gradient ${\partial e}/{\partial \mX^\ell}$ is required to back-propagate to the lower layer. Considering the total loss $e$ is the sum of the multinomial logistic loss $e_{\scriptscriptstyle a}$ and the regularization loss $e_{\scriptscriptstyle b}$, the gradient of $e$ with respect to $\mW^\ell$ is represented as  
\begin{equation}
\frac{\partial e}{\partial \mW^\ell} = \frac{\partial e_{\scriptscriptstyle a}}{\partial \mW^\ell} + \frac{\partial e_{\scriptscriptstyle b}}{\partial \mW^\ell},
\label{eq:gradientWeight1}
\end{equation}
and this is rewritten by the chain rule~\cite{deeplearning, PRML, chainrule}, as follows: 
\begin{equation}
\frac{\partial e}{\partial \mW^\ell} = \frac{\partial e_{\scriptscriptstyle a}}{\partial \mX^{\ell+1}} \frac{\partial \mX^{\ell+1}}{ \partial \mW^\ell} + \frac{\partial e_{\scriptscriptstyle b}}{\partial \mW^\ell}.  
\label{eq:chainrule1}
\end{equation}
For the shared weight $\mW_{t,r,u,v}^\ell$, the gradient of \eqref{eq:chainrule1} is expanded as 
\begin{equation}
\begin{split}
& \frac{\partial e}{\partial \mW^\ell_{t,r,u,v}} = \frac{\partial e_a}{\partial \mW^\ell_{t,r,u,v}} + \frac{\partial e_b}{\partial \mW^\ell_{t,r,u,v}} \\
& = \sum_m \sum_n \frac{\partial e_a}{\partial \mX^{\ell+1}_{t,m,n}} \frac{\partial \mX^{\ell+1}_{t,m,n}}{ \partial \mW^\ell_{t,r,u,v}} + \lambda \mW^\ell_{t,r,u,v}.
\label{eq:gradientWeight2}
\end{split}
\end{equation}
Recalling \eqref{eq:output1}, since an output node has the input nodes determined by the depth-adaptive receptive field, $\mS$ is required to decode the connections from input nodes to output nodes~(see \fref{fig:sparsity}). Considering this variation of receptive field, the second factor of the multinomial logistic loss $e_{\scriptscriptstyle a}$ is evaluated as
\begin{equation}
\frac{\partial \mX^{\ell+1}_{t,m,n}}{ \partial \mW^\ell_{t,r,u,v}} = \mX^\ell_{r,m+ \mS^\ell_{r,m,n} u,\, n+ \mS^\ell_{r,m,n} v}.
\end{equation}     
To compute the first factor of $e_{\scriptscriptstyle a}$, let's first consider a specific connection between the input node $({r, m+ \mS^\ell_{r,m,n} u,\, n + \mS^\ell_{r,m,n} v})$ and  the output node $(t,m,n)$. The gradient of 
this specific connection is back-propagated as follows:
\begin{equation}
\begin{split}
&\frac{\partial e_a}{\partial \mX^\ell_{r, m+ \mS^\ell_{r,m,n} u,\, n + \mS^\ell_{r,m,n} v}} \\
&= \frac{\partial e_a}{\partial \mX^{\ell+1}_{t,m,n}} \frac{\partial \mX^{\ell+1}_{t,m,n}}{\partial \mX^\ell_{r, m+ \mS^\ell_{r,m,n} u,\,  n + \mS^\ell_{r,m,n} v}} \\
&= \frac{\partial e_a}{\partial \mX^{\ell+1}_{t,m,n}} \mW^\ell_{t,r,u,v}.
\label{eq:gradientInput}
\end{split}
\end{equation}
In \eqref{eq:gradientInput}, the output node $\mX_{t,m,n}^{\ell+1}$ is influenced by the multiple input nodes,  then the gradient ${\partial e_a}/{\partial \mX^\ell}$ is computed by the iterative accumulations over the feature spaces and the spatial locations, as summarized in Algorithm~\ref{alg:gradient}.

\input{alg_1}

Finally, the weight matrix $\mW^\ell$ is updated using the stochastic gradient descent algorithm with momentum~\cite{deeplearning} because we use small batch of training data to compute the gradients. At an iteration $i$, suppose the current weight matrix is denoted as $\mW^{\ell,i}$, then, the weight matrix $\mW^{\ell,i+1}$ at the iteration $i+1$ is updated considering the previous update and the computed gradient as follows: 
\begin{equation}
\mW^{\ell,i+1} = \mW^{\ell,i} + \mu (\mW^{\ell,i} - \mW^{\ell,i-1}) - \gamma (\partial e / \partial \mW^{\ell,i})
\label{eq:weight_update}
\end{equation}
where $\mu$ and $\gamma$ denote the momentum and the learning rate, respectively. The momentum $\mu$ was chosen as $0.99$ for Frontend module and $0.9$ for DeepLab, and the learning rate is explained in \sref{sec:result}. 

%% file: alg_1.tex
\begin{algorithm}
  \caption{Gradient of loss with respect to input}
\label{alg:gradient}
\renewcommand{\algorithmicrequire}{\textbf{Input: }}
\renewcommand{\algorithmicensure}{\textbf{Output: }}
\begin{algorithmic}
\STATE{\algorithmicrequire{$\partial e_a/\partial \mX^{\ell+1}$, $\mW^\ell$, $\mS^\ell$}}
\STATE{\algorithmicensure{${\partial e_a}/{\partial \mX^\ell}$}}
\newline
\STATE{
initialize ${\partial e_a}/{\partial \mX^\ell} = 0$
}
\FOR{all $t, r, m, n, u, v$}
\STATE{
\begin{equation*}
\frac{\partial e_a}{\partial \mX^\ell_{r, m+ \mS^\ell_{r,m,n} u,\,  n + \mS^\ell_{r,m,n} v}} \mathrel{+}= \frac{\partial e_a}{\partial \mX^{\ell+1}_{t,m,n}} \mW^\ell_{t,r, u, v} 
\end{equation*}
}
\ENDFOR
\end{algorithmic}
\end{algorithm}

%% file: 3_6_proof.tex
\subsection{Proof of Depth-Invariance} \label{sec:proof}
In this section, we present the mathematical proof of the depth-invariance property of the proposed networks. We first simplify the convolution in~\eqref{eq:output1} by considering a single channel one-dimensional input and output. We, then, apply the proposed convolution to an input at different distances from the camera. By demonstrating that the outputs are equivalent regardless of the distances, we prove that the proposed DaM convolution is depth-invariant.

Considering a neural network having a single channel (feature space), \eqref{eq:output1} is substituted as follows:
\begin{equation}
\mX^{\ell+1}_{m,n} = f\bigg(\sum_{u} \sum_{v} {\mW_{u,v} \mX^\ell_{m+\mS_{m,n} u,\,  n+\mS_{m,n} v}} + b^{\ell} \bigg).
\label{eq:SingleActivation}
\end{equation}
For the one-dimensional input, \eqref{eq:SingleActivation} is further simplified as 
\begin{equation}
\vx_{m}^{\ell+1} = f\bigg( \sum_{u} \vw_{u}^{\ell} \vx_{m+ \vs_{m} u}^{\ell}+b^\ell \bigg).
\end{equation}

\input{fig_proof}

Let's first consider the example in~\fref{fig:proof}, showing the proposed convolution layers for the input at distance $d$ in \fref{fig:proof}(a) and at distance $d/2$ in \fref{fig:proof}(b). In the example, the size of kernel is set to $3$, and the size $\vs$ of receptive field is 1 at distance $d$. Then, the output $\vx^2_5$ of the first convolution layer in \fref{fig:proof}(a) is
\begin{equation}
\begin{split}
\vx^2_{5} &= f\Big( \sum_{u^1=-1}^{1} {\vw^1_{u^1} \vx^1_{5+ u^1}}+b^1 \Big)\\
&= f\Big(\vw_{-1}^1 \vx^1_{4} + \vw^1_{0} \vx^1_{5} + \vw^1_{1} \vx^1_{6}+b^1 \Big), 
\end{split}
\end{equation}
and the output $\vx^3_5$ of the second convolution layer is 
\begin{equation}
\begin{split}
\vx^3_{5}=& f\Big(\vw_{-1}^2 \vx^2_{4} + \vw_{0}^2 \vx^2_{5} + \vw_{1}^2 \vx^2_{6} + b^2 \Big)\\
=& f\Big( \vw_{-1}^2 \cdot f(\vw_{-1}^1 \vx_{3} + \vw_{0}^1 \vx_{4} + \vw_{1}^1 \vx_{5} + b^1) \Big.\\
&+ \vw_{0}^2 \cdot f(\vw_{-1}^1 \vx_{4} + \vw_{0}^1 \vx_{5} + \vw_{1}^1 \vx_{6} + b^1) \\ 
& \Big.+ \vw_{1}^2 \cdot f(\vw_{-1}^1 \vx_{5} + \vw_{0}^1 \vx_{6} + \vw_{1}^1 \vx_{7} + b^1) +b^2 \Big). 
\end{split}
\end{equation}
In \fref{fig:proof}(b), the distance from the camera decreases to $d/2$, thus the size of the object on an image plane is doubled comparing to the size at $d$~(see \fref{fig:pinholeCamera}). Let $\vxh$ denote the input at distance $d/2$ and suppose $\vxh_5$ corresponds to $\vx_5$. Then, $\vxh_{5+2v}$ is equivalent to $\vx_{5+v}$ for $\forall v \in \Z$~(e.g. $\vx_6 = \vxh_7$ for $v = 1$). Since its receptive field increases by a factor of 2 by the relation of \eqref{eq:scale}, the output $\vxh^2_5$ of the first convolution layer is consequently equivalent to $\vx^2_{5}$: 
\begin{equation}
\begin{split}
\vxh^2_{5} &= f\Big(\sum_{u^1=-1}^{1} {\vw_{u^1}^{1} \vxh^1_{5+ 2 u^1}} + b^1 \Big)\\
&= f\Big(\vw_{-1}^1 \vxh^1_{3} + \vw_{0}^1 \vxh^1_{5} + \vw_{1}^1 \vxh^1_{7} + b^1\Big)\\
&= f\Big(\vw_{-1}^1 \vx_{4} + \vw_{0}^1 \vx_{5} + \vw_{1}^1 \vx_{6} + b^1\Big)\\
&= \vx^2_{5},
\end{split}
\label{eq:example1}
\end{equation}
and the output $\vxh^3_5$ of the second convolution layer is
\begin{equation}
\begin{split}
\vxh^3_{5} =& f \Big(\vw_{-1}^2 \vxh^2_{3} + \vw_{0}^2 \vxh^2_{5} + \vw_{1}^2 \vxh^2_{7} + b^2 \Big)\\
=& f \Big(\vw_{-1}^2 \cdot f(\vw_{-1}^1 \vxh^1_{1} + \vw_{0}^1 \vxh^1_{3} + \vw_{1}^1 \vxh^1_{5} + b^1) \Big.\\
    &+ \vw_{0}^2 \cdot f(\vw_{-1}^1 \vxh^1_{3} + \vw_{0}^1 \vxh^1_{5} + \vw_{1}^1 \vxh^1_{7} + b^1) \\ 
    & \Big. + \vw_{1}^2 \cdot f(\vw_{-1}^1 \vxh^1_{5} + \vw_{0}^1 \vxh^1_{7} + \vw_{1}^1 \vxh^1_{9} + b^1) + b^2 \Big)\\
=& f \Big(\vw_{-1}^2 \cdot f(\vw_{-1}^1 \vx^1_{3} + \vw_{0}^1 \vx^1_{4} + \vw_{1}^1 \vx^1_{5} + b^1)\Big.\\
    &+ \vw_{0}^2 \cdot f(\vw_{-1}^1 \vx^1_{4} + \vw_{0}^1 \vx^1_{5} + \vw_{1}^1 \vx^1_{6} + b^1) \\ 
    &\Big. + \vw_{1}^2 \cdot f(\vw_{-1}^1 \vx^1_{5} + \vw_{0}^1 \vx^1_{6} + \vw_{1}^1 \vx^1_{7} + b^1)+ b^2 \Big) \\
    =&\vx^3_{5}.
\end{split}
\label{eq:example2}
\end{equation}
We conclude from this simple example that the proposed convolution extracts depth-invariant activations.

From the fact that $\vxh^{\ell}_{m+\vsh^{\ell}_{m} u^\ell}$ is equivalent to $\vx^{\ell}_{m+\vs^{\ell}_{m} u^\ell}$, the demonstration of depth-invariant activations is generalized as 
\begin{equation}
\begin{split}
&\widehat{\mX}^{\ell+1}_{t^\ell, m, n} \vspace{0.2cm}\\ 
&= f \bigg( \sum\limits_{r^\ell} \sum\limits_{u^\ell} \sum\limits_{v^\ell} {\mW^{\ell}_{t^\ell, r^\ell, u^\ell, v^\ell} \widehat{\mX}^{\ell}_{r^\ell, m+\widehat{\mS}^{\ell}_{m,n} u^\ell, n+\widehat{\mS}^{\ell}_{m,n} v^\ell}} + b^\ell \bigg) \vspace{0.2cm} \\
&= f \bigg( \sum\limits_{r^\ell} \sum\limits_{u^\ell} \sum\limits_{v^\ell} \mW^{\ell}_{t^\ell, r^\ell, u^\ell, v^\ell} \widehat{\mX}^{\ell}_{r^\ell, m+(\mS^\ell_{m,n}/g) u^\ell, n+(\mS^\ell_{m,n}/g) v^\ell} \\
& \quad+ b^\ell \bigg) \vspace{0.2cm}\\
&= f \bigg( \sum\limits_{r^\ell} \sum\limits_{u^\ell} \sum\limits_{v^\ell} \mW^{\ell}_{t^\ell, r^\ell, u^\ell, v^\ell} \mX^{\ell}_{r^\ell,m+\mS^{\ell}_{m,n} u^\ell, n+\mS^{\ell}_{m,n} v^\ell} + b^\ell \bigg) \vspace{0.2cm}\\
&= \mX^{\ell+1}_{t^\ell,m,n},
\end{split}
\label{eq:generalization}
\end{equation}
where $g$ is the ratio of distances between $\vxh$ and $\vx$. From the example of \eqref{eq:example1}, \eqref{eq:example2} and the generalization of \eqref{eq:generalization}, we conclude that the proposed convolution extracts depth-invariant activations by adjusting the size of receptive field.

%% file: fig_proof.tex
\begin{figure}[!t] \begin{center}
\begin{minipage}{0.45\linewidth}
\centerline{\includegraphics[scale=0.28]{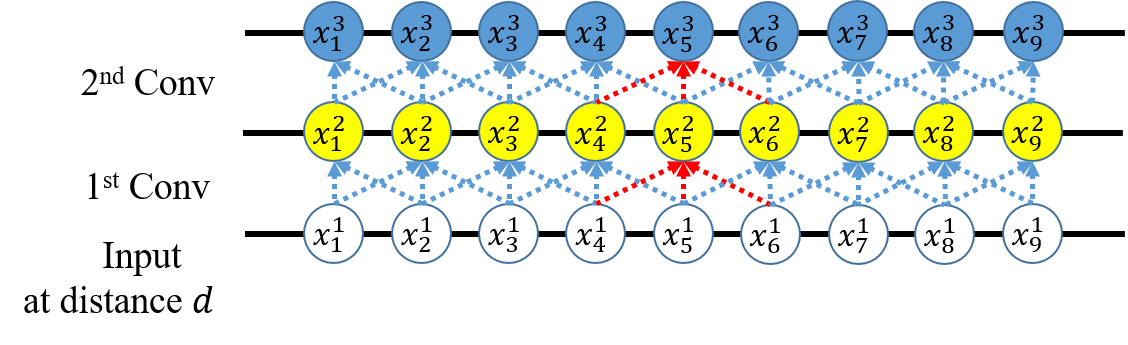}} \vspace{-0.2cm} 
\centerline{\footnotesize (a): Input at distance $d$} 
\end{minipage}
\vspace{0.4cm} \\ 
\begin{minipage}{0.45\linewidth}
\centerline{\includegraphics[scale=0.28]{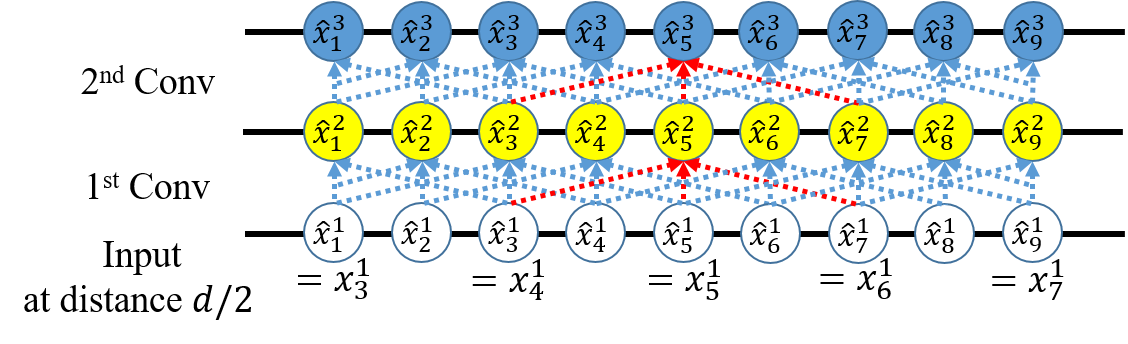}}
\centerline{\footnotesize (b): Input at distance $d/2$} 
\end{minipage}
   \caption{Example of the proposed DaM convolution layers at different distances to demonstrate depth-invariance.}
\label{fig:proof}
\end{center}\end{figure}

%% file: 4_experiments_results.tex
\section{Experiments and Results} \label{sec:result}
The proposed neural networks were tested on two applications: indoor semantic segmentation and hand segmentation for hand-object interaction. The experimental results verify that the proposed neural networks outperform original Frontend module~\cite{YuKoltun2016} and DeepLab~\cite{Chen16} without any additional layer or pre/post-processing. 

For comparison, we report pixel-wise accuracy, mean accuracy, mean intersection over union~(IoU), and frequency weighted~(FW) IoU for both experiments. Additionally, for hand segmentation, we report precision, recall, and $F_1$ score. Let $n_{ij}$ be the number of pixels which belong to the class $i$ and are predicted to the class $j$, and $c_t$ be the total number of classes. 
\begin{equation}
\begin{split}
&\text{Pixel accuracy} = \frac{\sum\limits_i n_{ii}}{\sum\limits_i \sum\limits_j n_{ij}},                                 \\
&\text{Mean accuracy} = \frac{1}{c_t}\sum\limits_i \bigg(\frac{n_{ii}}{\sum\limits_j n_{ij}}\bigg),                        \\
&\text{Mean IoU} = \frac{1}{c_t} \sum\limits_i \bigg( \frac{n_{ii}}{\sum\limits_j n_{ij} + \sum\limits_j n_{ji} - n_{ii}} \bigg),         \\
&\text{FW IoU} = \frac{1}{\sum\limits_i\sum\limits_j n_{ij}} \sum\limits_i \bigg(\frac{\sum\limits_j n_{ij} n_{ii}}{\sum\limits_j n_{ij} + \sum\limits_j n_{ji} - n_{ii}}\bigg), \\
&\text{Precision} = \frac{n_{11}}{n_{11} + n_{01}},    \\
&\text{Recall} = \frac{n_{11}}{n_{11} + n_{10}},          \\
&F_1 = \frac{2 n_{11}}{2 n_{11} + n_{01} + n_{10}},
\end{split}
\end{equation}
where for hand segmentation, class $1$ is hand, and class $0$ is others.

\input{tbl_result_nyud}

\subsection{Indoor Semantic Segmentation~(NYUDv2)}
\subsubsection{Dataset}  
The NYUDv2 dataset consists of 1,449 pairs of RGB-D images including various indoor scenes with pixel-wise annotations~\cite{nyudv2}. The pixel-wise annotations were coalesced into 40 dominant object categories by Gupta \etal~\cite{gupta2013}. We experimented with this 40 classes problem with the standard separation~\cite{nyudv2, gupta2013} of 795 training images and 654 testing images. 

\input{fig_result_nyud}

\subsubsection{Experiments}  
All the models were initialized using the VGG-16 model~\cite{vgg} trained using the ImageNet ILSVRC-2014 dataset~\cite{ILSVRC15} except for the input of RGB-HHA. Then, the models were fine-tuned using the NYUDv2 training dataset~\cite{nyudv2}. For the input of RGB-HHA, we initialized the model using the two fine-tuned models using NYUDv2 dataset~(one model using RGB images and the other model using HHA images). Then, we fine-tuned the model using the pair of RGB images and HHA images similar to~\cite{longcvpr, longpami}.
The initial base learning rate was selected by trying several learning rates~($\gamma$) with a factor of 10 such as $[10^{-9}, 10^{-10}, 10^{-11}, ... ]$. The decay factor~($\lambda$) of the weight matrix is chosen as $0.0005$. The models used in the experiments were selected based on the mean IoU score. During training, we computed the mean IoU score at every 1,000 iterations for the input of RGB-HHA and at every 2,000 iterations for the other inputs. 

For Frontend module, we used the multinomial logistic loss without normalization during training. So, the normalization term $1/h^{\ell} w^{\ell}$ was removed from (\ref{eq:loss}). The initial base learning rate was selected as $10^{-12}$ for the input of RGB-HHA and $10^{-10}$ for the other inputs. The scaling parameter $s_r^1$ for the first layer was set to $ \{1, 1.5, 2\}$ and $s_r$ for other layers was set to $\{0.5, 0.75, 1.0, 1.25\}$ for color images and depth maps and $\{0.75, 1.0, 1.25, 1.5\}$ for HHA images. If the mean IoU score stops improving, the base learning rate was decreased by a factor of 10. The training was terminated if the improvement of the score is negligible~($< 0.001$) or the score is not improved.  

For DeepLab, the initial base learning rate was selected as $10^{-4}$ for HHA images and $10^{-3}$ for other inputs. The learning rate was decreased using polynomial decay with the power of $0.9$ and the maximum iteration of $40,000$. The scaling parameters $s_r$ for all layers and for all inputs were set to be linearly distributed in $[0, 1.5)$.

\subsubsection{Results}  
We adopted the experimental settings in~\cite{longcvpr, longpami}. We considered the inputs of an RGB image, the concatenated image of an RGB image and a depth map~(early fusion), and an HHA encoded image~\cite{Gupta2014}. We also experimented with combining the scores from an RGB image and from an HHA encoded image~\cite{Gupta2014} at the last layer~(late fusion). Table~\ref{tab:nyudv2_result} and Fig.~\ref{fig:nyudv2_result} show the quantitative results and the qualitative results. The proposed method achieves the improvements without any additional layers or pre/post-processing. 

\input{tbl_multiscale}

\input{tbl_dam}

\input{tbl_multiscale_eval}
\input{tbl_receptive}

\begin{figure}[t] \begin{center}
\begin{minipage}{0.95\linewidth}
\centerline{\includegraphics[scale=0.33]{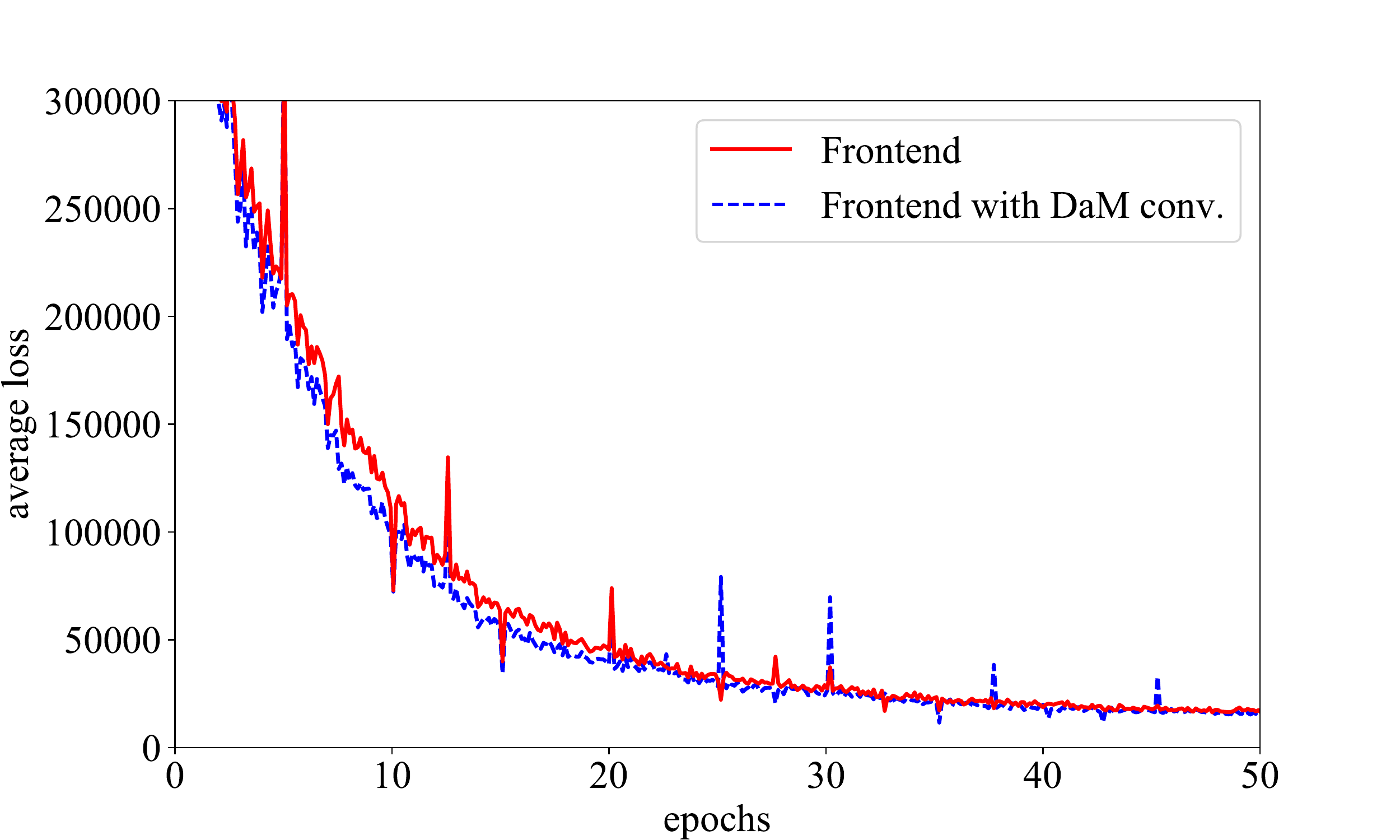}}
\end{minipage}
\end{center}
   \caption{The comparison of convergence curves between Frontend module~\cite{YuKoltun2016} and the network with the proposed DaM convolution for the input of RGB.}
\label{fig:convergence}
\end{figure}

\subsubsection{Analysis}  
We experimentally analyze the effects of multiscale parameters $s^l_r$ in \tref{tab:multiscale_result}. The analysis shows that the proposed method outperforms other methods using the parameters in the reasonable ranges. We also analyze the effects of applying the different number of the DaM convolution in \tref{tab:dam}. The experiments demonstrate that replacing all convolution layers outperforms other settings. The processing time is measured using a machine with Intel i7-4790K CPU and Nvidia Tesla K40c. \tref{tab:random_multiscale_evaluation} shows that multi/random scale evaluation has the chance of further improving the segmentation performance. In the multiscale evaluation, the final results are combined with the results of original, twice enlarged, and half-scaled inputs. In the random scale evaluation, the final results are fused from the results of original and two randomly scaled inputs. The results using both scaling are combined with the results of the previously mentioned five inputs. \tref{tab:dilation_analysis} demonstrates that simply increasing receptive fields in DeepLab does not improve the accuracy. Lastly, we show the convergence curve for Frontend module~\cite{YuKoltun2016} and the network with the proposed DaM convolution in \fref{fig:convergence}. The average loss is computed using the losses from 100 iterations. The graph shows that the proposed method converges slightly faster than Frontend module. 

\input{fig_dataset_hand}

\input{fig_dataset_dist_hand}

\subsection{Hand-Object Interaction~(HOI)} \label{sec:HOI}
\subsubsection{Dataset}  
We collected a new dataset\footnote{\url{https://github.com/byeongkeun-kang/HOI-dataset}} using Microsoft Kinect v2 since we were not able to find a publicly available dataset for hand-object interaction with pixel-wise annotation. The collected dataset consists of more than 9,175 pairs of depth maps and color images from 6 people~(3 males and 3 females) interacting with 21 different objects. In addition, the dataset includes the cases of one hand and both hands in a scene. Ground truth was labeled by wearing a color glove during data collection and by finding the color of the glove on the color images.

\input{tbl_result_hand}
\input{fig_result_hand}

To increase the variation of the dataset further~(e.g. the distance from the camera to hands), 18,350 pairs of images were augmented by moving the camera closer/further to/from the scene as shown in Fig.~\ref{fig:AugDataset}. In total, the augmented dataset has $27,525$ pairs of depth maps and ground truth labels. Indeed, the standard deviation of the augmented data increases to $725$ relative to that of the collected dataset is $225$, as evidenced in \fref{fig:distribution}(a). The distances of the augmented dataset are distributed at more diverse distances as demonstrated in \fref{fig:distribution}(b).

Among $27,525$ pairs, we used 19,470 pairs for training, 2,706 pairs for validation, and 5,349 pairs for testing.

\subsubsection{Experiments}  
All the models were initialized using the VGG-16 model~\cite{vgg} that were trained using the ImageNet ILSVRC-2014 dataset~\cite{ILSVRC15}. Then, the initial models were fine-tuned using the HOI training dataset. The initial base learning rate $\gamma$ was selected by trying several learning rates with the factor of 10 such as $[10^{-3}, 10^{-4}, 10^{-5}, ... ]$. In most cases, the initial learning rate was selected as $10^{-4}$. The decay factor $\lambda$ of the weight matrix is chosen as $0.0005$.

The models used in the experiments were selected based on the $F_1$ score on the validation dataset. During training, we computed the $F_1$ score on the validation dataset at every 4,000 iterations. If the $F_1$ score stops improving, the base learning rate was decreased by a factor of 10. The training was terminated if the improvement of the score is negligible~($< 0.001$) or the score is not improved. 
The multiscale parameter $s_r^1$ was set to $\{1, 1.5, 2\}$ and $s_r$ for other layers was set to $\{0.75, 1.0, 1.25, 1.5\}$ for each quarter of the feature spaces in each convolution layer.

\subsubsection{Results}  
The performance of the proposed methods and the comparing methods is tabulated in \tref{tab:resultHOI} for the inputs of the depth maps and the HHA encoded images~\cite{Gupta2014}. The visual segmentation results are displayed in \fref{fig:result}. The proposed neural network improves about $14\%$~(depth maps) and $3\%$~(HHA) in $F_1$ score relative to the baseline Frontend model~\cite{YuKoltun2016}. Moreover, the proposed network with the input of depth map achieves higher $F_1$ score and mean IoU than Frontend module with the input of the HHA encoded image. These results verify that the proposed networks improve segmentation performance without any additional layer or pre/post-processing.

%% file: tbl_result_nyud.tex
\begin{table*}
\begin{center}
\caption{The quantitative results of the NYUDv2 dataset. The scores are scaled by a factor of 100. Bold face and blue color emphasize the best performance.}\label{tab:nyudv2_result}
\renewcommand{\arraystretch}{1.2} 
\begin{tabu}{X[c]|X[c]|X[c]||X[c]|X[c]|X[c]|X[c]} 
\hline
\multirow{2}{*}{Input} & \multicolumn{2}{c||}{Method} & \multirow{2}{*}{Pixel accu.} & \multirow{2}{*}{Mean accu.} & \multirow{2}{*}{FW IoU} & \multirow{2}{*}{Mean IoU} \\
\cline{2-3}
  & Architecture & DaM conv. &  &  &  & \\
\cline{2-4}
\hline\hline
\multicolumn{3}{c||}{Gupta \etal~\cite{Gupta2014}} & 60.3 & - & 47.0 & 28.6 \\
\hline
\multirow{8}{*}{RGB} & FCN-32s~\cite{longcvpr}	& -	& 60.0 & 42.2 & 43.9 & 29.2 \\
			& FCN-32s~\cite{longpami} 			& -	& 61.8 & 44.7 & 46.0 & 31.6 \\
			& FCN-16s~\cite{longpami} 			& -	& 62.3 & 45.1 & 46.8 & 32.0 \\
			& FCN-8s~\cite{longpami} 			& -	& 62.1 & 46.1 & 47.2 & 32.4 \\
\cline{2-7}
 			& \multirow{2}{*}{Frontend~\cite{YuKoltun2016}} 	& -	& 62.1 & 45.8 & 46.6 & 32.3 \\
 			&  						& \checkmark & \bb{63.7} & \bb{47.2} & \bb{48.3} & \bb{33.3} \\
\cline{2-7}
	             & \multirow{2}{*}{DeepLab~\cite{Chen16}} & - & 63.8 & 46.2 & 48.3 & 33.7 \\
       	      & 						& \checkmark & \bb{64.3} & \bb{47.3} & \bb{49.0} & \bb{34.3} \\
\hline
\multirow{8}{*}{RGB-D} & FCN-32s~\cite{longcvpr}	& -	& 61.5 & 42.4 & 45.5 & 30.5\\ 
			 & FCN-32s~\cite{longpami} 			& -	& 62.1 & 44.8 & 46.3 & 31.7\\ 
			 & FCN-16s~\cite{longpami} 			& -	& 62.3 & 45.4 & 46.8 & 32.2\\ 
			 & FCN-8s~\cite{longpami} 			& -	& 62.7 & 46.0 & 47.4 & 32.5\\ 
\cline{2-7}
			 & \multirow{2}{*}{Frontend~\cite{YuKoltun2016}}		& -	& 62.1 & 46.2 & 46.8 & 32.5 \\
			 &  				& \checkmark	& \bb{63.8} & \bb{47.1} & \bb{48.3} & \bb{33.2} \\
\cline{2-7}
	             & \multirow{2}{*}{DeepLab~\cite{Chen16}} & - & 63.7 & \bb{47.2} & 48.3 & 33.3 \\
             	& 				& \checkmark & \bb{64.7} & 47.0 & \bb{49.4} & \bb{34.4} \\
\hline
\multirow{8}{*}{HHA~\cite{Gupta2014}} & FCN-32s~\cite{longcvpr}	& -	& 57.1 & 35.2 & 40.4 & 24.2 \\
			& FCN-32s~\cite{longpami} 			& -	& 58.3 & 35.7 & 41.7 & 25.2 \\
			& FCN-16s~\cite{longpami} 			& -	& 57.5 & 36.0 & 41.7 & 25.3 \\
			& FCN-8s~\cite{longpami} 			& -	& 56.8 & 36.7 & 41.9 & 25.6 \\
\cline{2-7}
			& \multirow{2}{*}{Frontend~\cite{YuKoltun2016}}	& -	& 56.7 & \bb{38.5} & 41.8 & 25.9 \\
			&  				 & \checkmark	& \bb{58.2} & 38.4 & \bb{42.6} & \bb{26.4} \\
\cline{2-7}
             	& \multirow{2}{*}{DeepLab~\cite{Chen16}} 	& - 	& 57.9 & \bb{40.0} & 42.6 & 26.9 \\
             	& 				& \checkmark & \bb{60.6} & 38.3 & \bb{44.1} & \bb{27.7} \\
\hline
\multirow{9}{*}{RGB-HHA} & FCN-32s~\cite{longcvpr} 	& -	& 64.3 & 44.9 & 48.0 & 32.8 \\
			& FCN-32s~\cite{longpami} 			& -	& 65.3 & 44.0 & 48.6 & 33.3 \\
			& FCN-16s~\cite{longcvpr} 			& -	& 65.4 & 46.1 & 49.5 & 34.0 \\
			& FCN-16s~\cite{longpami} 			& -	& 67.0 & 47.2 & 51.1 & 35.8 \\
			& FCN-8s~\cite{longpami} 			& -	& 66.8 & 47.8 & 51.4 & 36.1 \\
\cline{2-7}
 			& \multirow{2}{*}{Frontend~\cite{YuKoltun2016}} 	& -	& 66.6 & 48.1 & 51.0 & 36.0 \\
			&				& \checkmark		& \bb{67.5} & \bb{48.9} & \bb{51.9} & \bb{36.8} \\
\cline{2-7}
             	& \multirow{2}{*}{DeepLab~\cite{Chen16}} & - & 66.9 & \bb{49.6} & 51.5 & 37.0 \\
             	& 				& \checkmark 		& \bb{68.4} & 49.0 & \bb{52.8} & \bb{37.6} \\			
\hline
\end{tabu}
\end{center}
\end{table*}

%% file: fig_result_nyud.tex
\begin{figure*}[!t] \begin{center}
\begin{minipage}{0.15\linewidth}
\centerline{\includegraphics[scale=0.13]{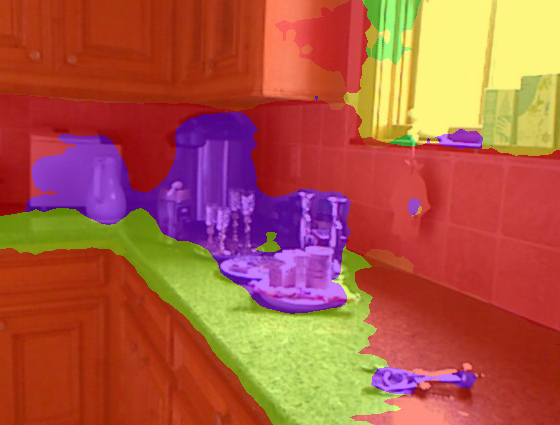}}
\end{minipage}
\begin{minipage}{0.15\linewidth}
\centerline{\includegraphics[scale=0.13]{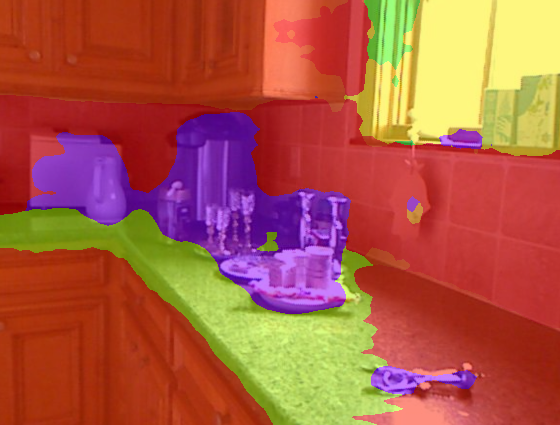}}
\end{minipage}
\begin{minipage}{0.15\linewidth}
\centerline{\includegraphics[scale=0.13]{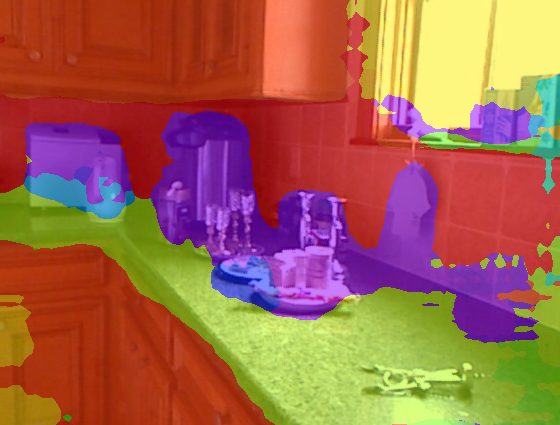}}
\end{minipage}
\begin{minipage}{0.15\linewidth}
\centerline{\includegraphics[scale=0.13]{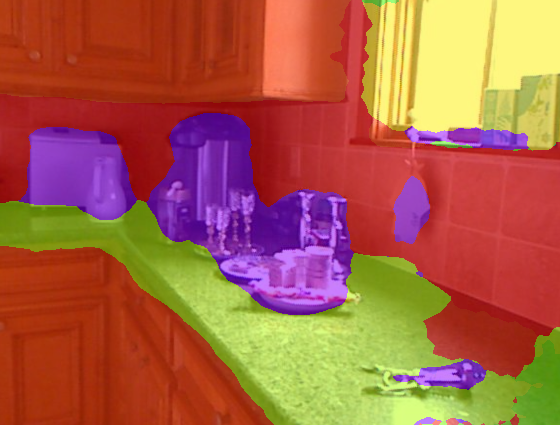}}
\end{minipage}
\begin{minipage}{0.15\linewidth}
\centerline{\includegraphics[scale=0.13]{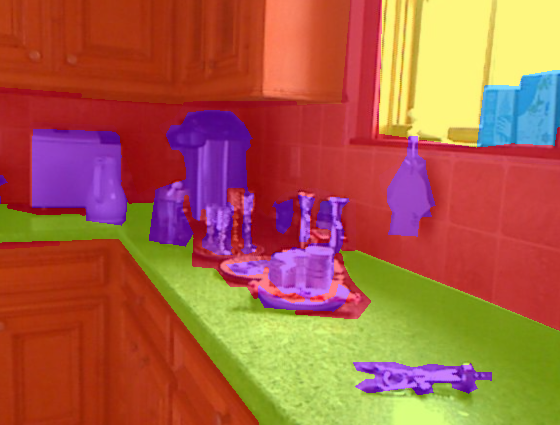}}
\end{minipage}
\\
\vspace{0.1cm}
\begin{minipage}{0.15\linewidth}
\centerline{\includegraphics[scale=0.13]{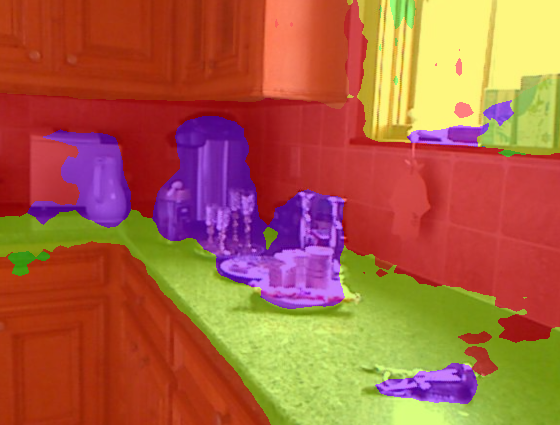}}
\end{minipage}
\begin{minipage}{0.15\linewidth}
\centerline{\includegraphics[scale=0.13]{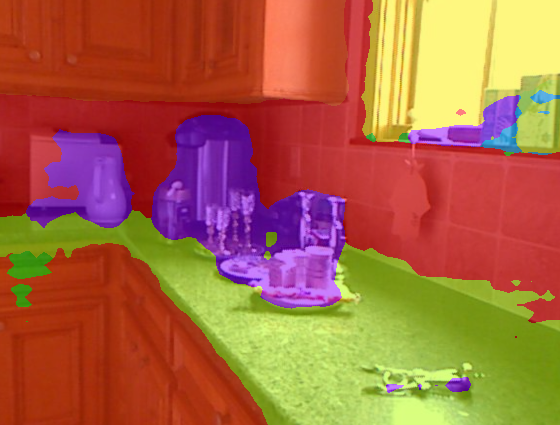}}
\end{minipage}
\begin{minipage}{0.15\linewidth}
\centerline{\includegraphics[scale=0.13]{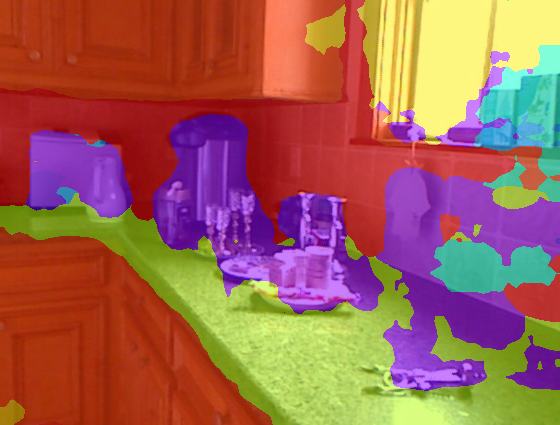}}
\end{minipage}
\begin{minipage}{0.15\linewidth}
\centerline{\includegraphics[scale=0.13]{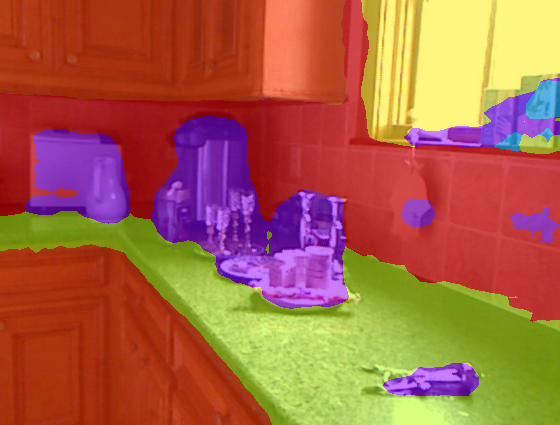}}
\end{minipage}
\begin{minipage}{0.15\linewidth}
\centerline{\includegraphics[scale=0.13]{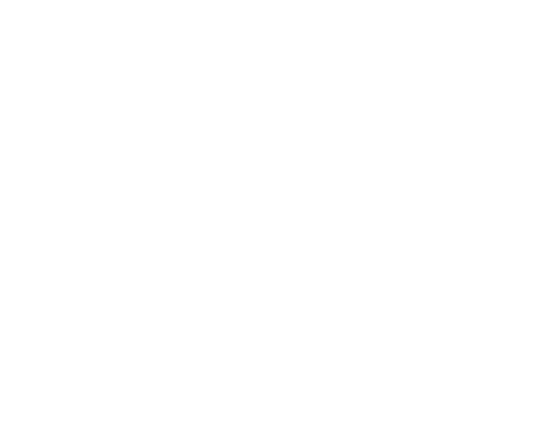}}
\end{minipage}
\\
\vspace{0.1cm}
\line(1,0){400}

\vspace{0.1cm}
\begin{minipage}{0.15\linewidth}
\centerline{\includegraphics[scale=0.13]{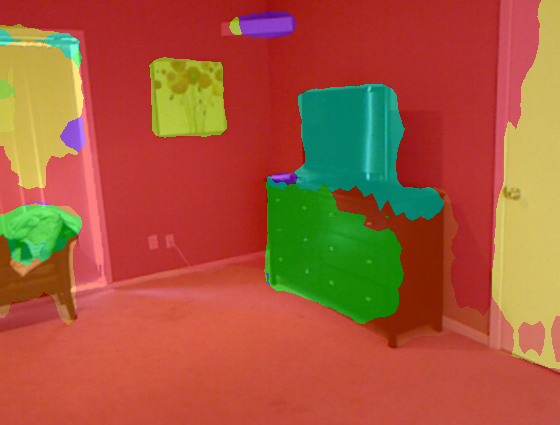}}
\end{minipage}
\begin{minipage}{0.15\linewidth}
\centerline{\includegraphics[scale=0.13]{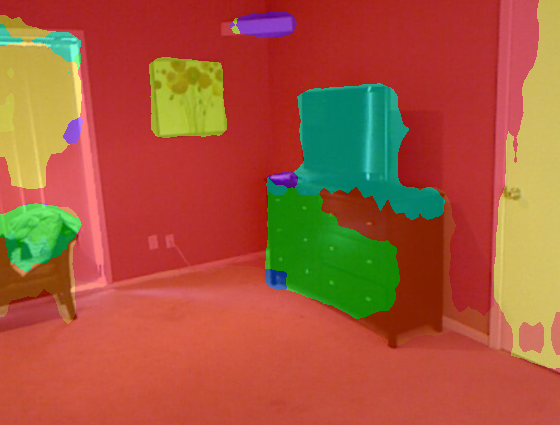}}
\end{minipage}
\begin{minipage}{0.15\linewidth}
\centerline{\includegraphics[scale=0.13]{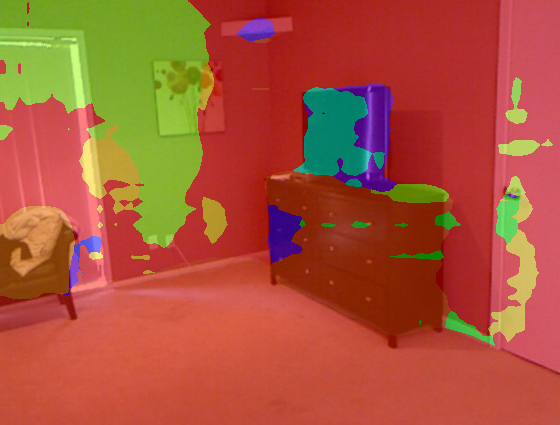}}
\end{minipage}
\begin{minipage}{0.15\linewidth}
\centerline{\includegraphics[scale=0.13]{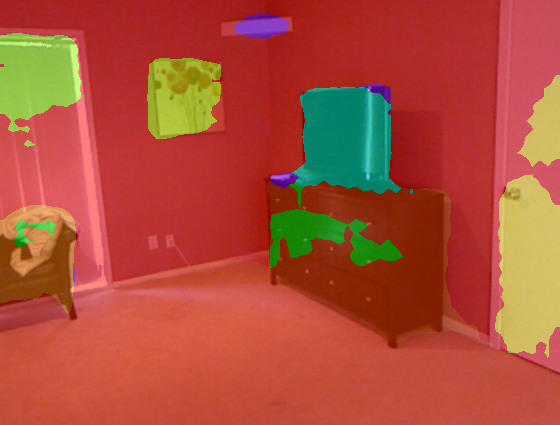}}
\end{minipage}
\begin{minipage}{0.15\linewidth}
\centerline{\includegraphics[scale=0.13]{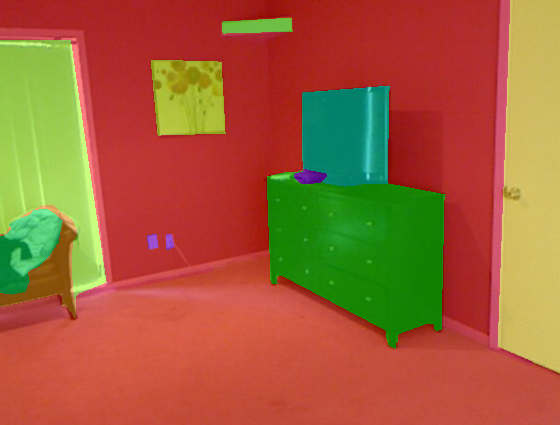}}
\end{minipage}
\\
\vspace{0.1cm}
\begin{minipage}{0.15\linewidth}
\centerline{\includegraphics[scale=0.13]{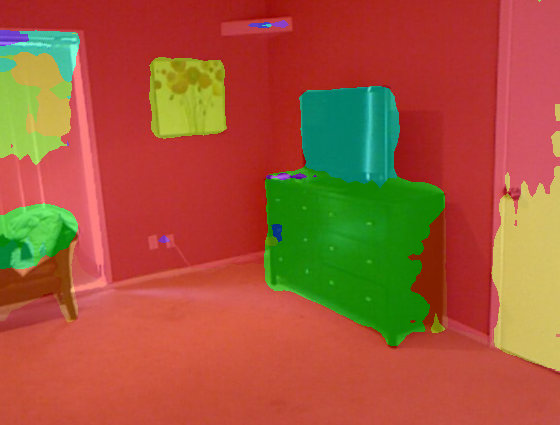}}
\end{minipage}
\begin{minipage}{0.15\linewidth}
\centerline{\includegraphics[scale=0.13]{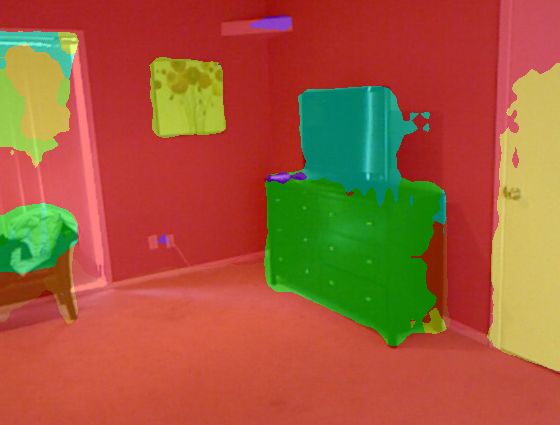}}
\end{minipage}
\begin{minipage}{0.15\linewidth}
\centerline{\includegraphics[scale=0.13]{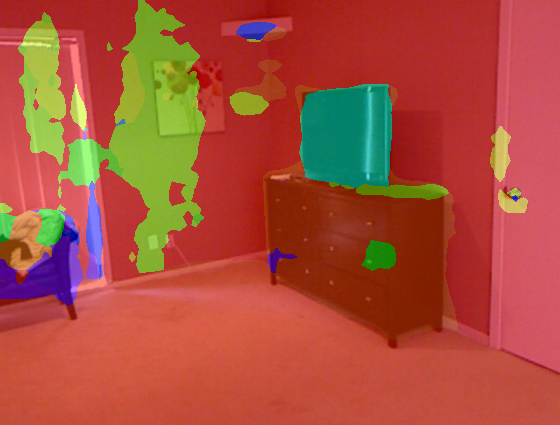}}
\end{minipage}
\begin{minipage}{0.15\linewidth}
\centerline{\includegraphics[scale=0.13]{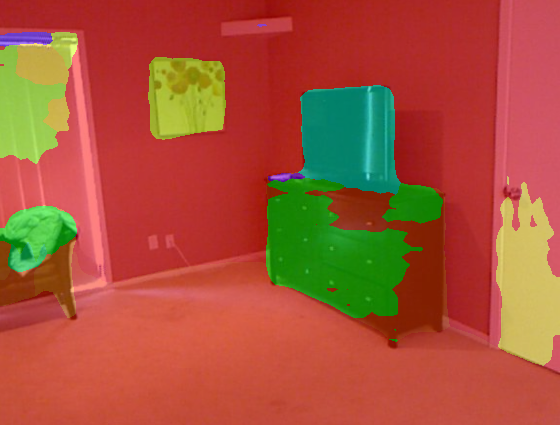}}
\end{minipage}
\begin{minipage}{0.15\linewidth}
\centerline{\includegraphics[scale=0.13]{img/NYUD_dataset/empty.jpg}}
\end{minipage}
\\
\vspace{0.1cm}
\line(1,0){400}

\vspace{0.1cm}
\begin{minipage}{0.15\linewidth}
\centerline{\includegraphics[scale=0.13]{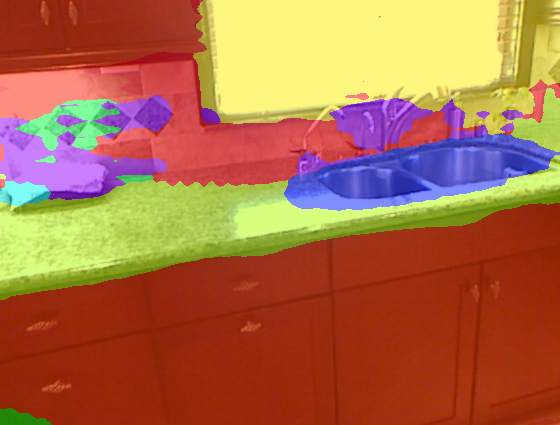}}
\end{minipage}
\begin{minipage}{0.15\linewidth}
\centerline{\includegraphics[scale=0.13]{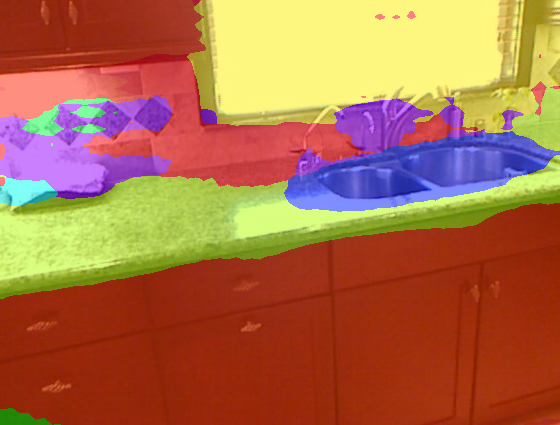}}
\end{minipage}
\begin{minipage}{0.15\linewidth}
\centerline{\includegraphics[scale=0.13]{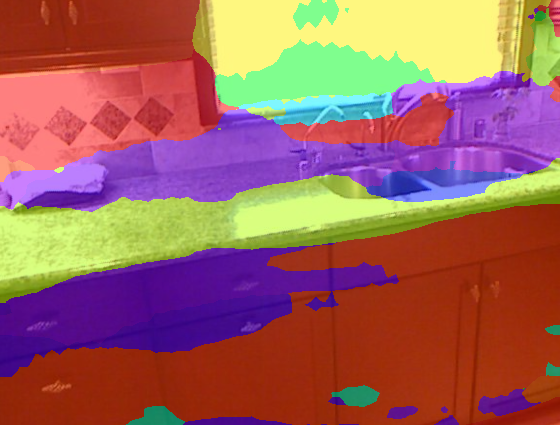}}
\end{minipage}
\begin{minipage}{0.15\linewidth}
\centerline{\includegraphics[scale=0.13]{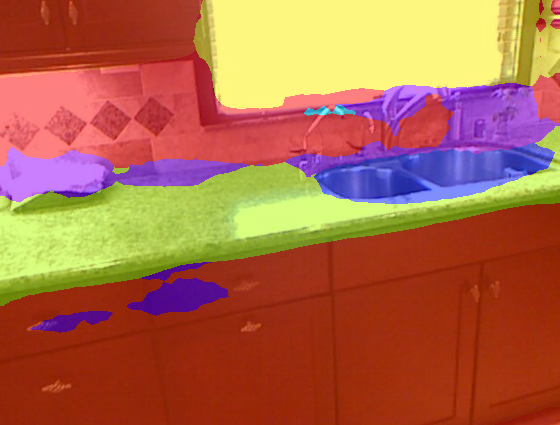}}
\end{minipage}
\begin{minipage}{0.15\linewidth}
\centerline{\includegraphics[scale=0.13]{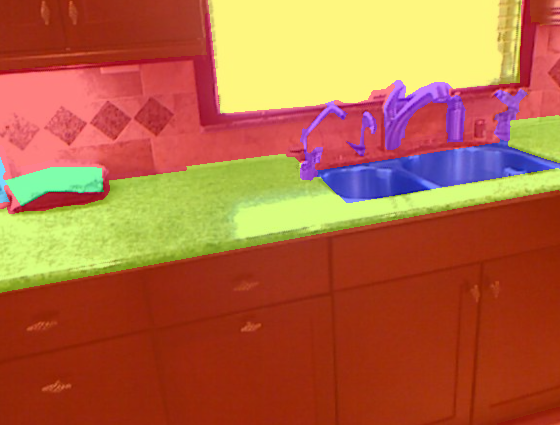}}
\end{minipage}
\\
\vspace{0.1cm}
\begin{minipage}{0.15\linewidth}
\centerline{\includegraphics[scale=0.13]{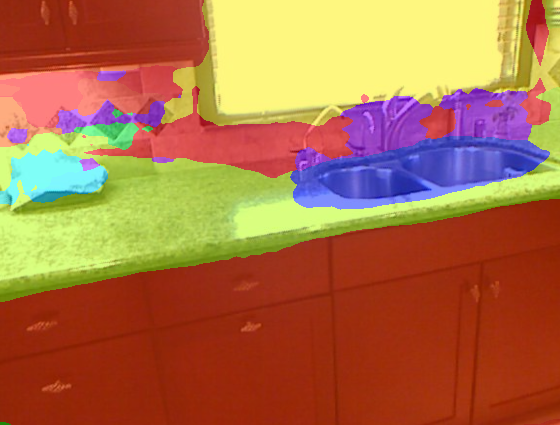}}
\centerline{\footnotesize (a): Input of RGB} 
\end{minipage}
\begin{minipage}{0.15\linewidth}
\centerline{\includegraphics[scale=0.13]{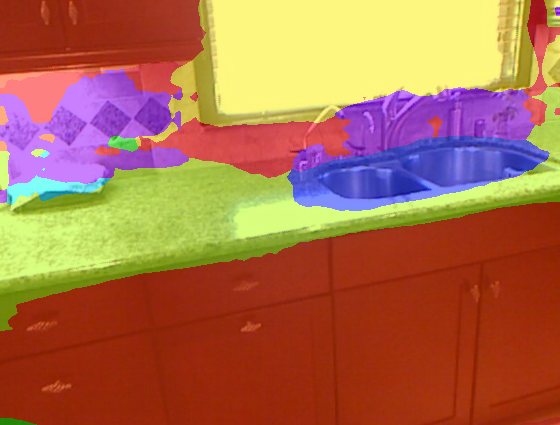}}
\centerline{\footnotesize (b): Input of RGB-D} 
\end{minipage}
\begin{minipage}{0.15\linewidth}
\centerline{\includegraphics[scale=0.13]{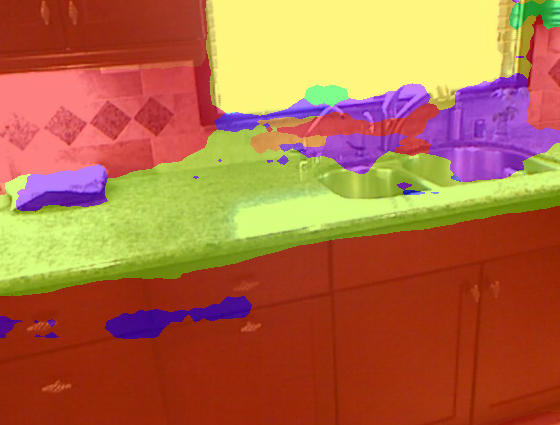}}
\centerline{\footnotesize (c): Input of HHA} 
\end{minipage}
\begin{minipage}{0.15\linewidth}
\centerline{\includegraphics[scale=0.13]{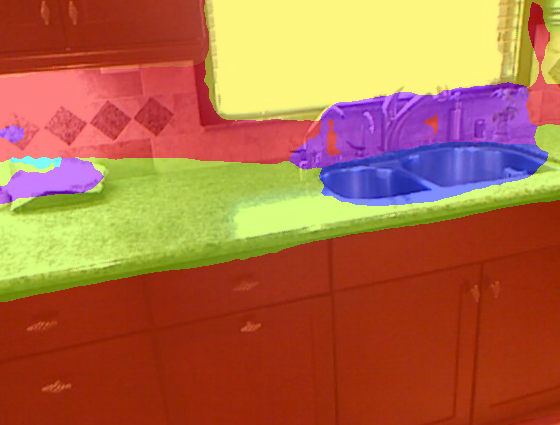}}
\centerline{\footnotesize (d): Input of RGB-HHA} 
\end{minipage}
\begin{minipage}{0.15\linewidth}
\centerline{\includegraphics[scale=0.13]{img/NYUD_dataset/empty.jpg}}
\centerline{\footnotesize (e): Ground truth labels} 
\end{minipage}
   \caption{The qualitative comparison of the result for the NYUDv2 dataset. The odd rows show the results of Frontend module~\cite{YuKoltun2016}, and the even rows show the results of the network with the proposed DaM convolution. The results and the ground truth labels are visualized on the color images for better visualization.}
\label{fig:nyudv2_result}
\end{center}\end{figure*}

%% file: tbl_multiscale.tex
\begin{table}
\begin{center}
\caption{Ablation study of selecting multiscale paramter $s_r$ for Frontend module.}\label{tab:multiscale_result}
\renewcommand{\arraystretch}{1.2} 
\begin{tabu}{c|c||X[c]|X[c]|X[c]|X[c]} 
\hline
\multicolumn{2}{c||}{Multiscale parameter $s_r$}  & Pixel & Mean & FW & Mean\\
\cline{1-2}
First conv. & Other conv.& accu. & accu. & IoU & IoU\\
\hline\hline
$[1, 1.25, 1.5]$	& $[0.5, 0.75, 1, 1.25]$	& 63.6 & 46.9 & 48.1 & 33.1 \\
$[1, 1.5, 2]$		& $[0.25, 0.5, 0.75, 1]$	& \bb{63.7} & 46.2 & 48.2 & 32.9 \\
$[1, 1.5, 2]$ 	& $[0.5, 0.75, 1, 1.25]$	& \bb{63.7} & \bb{47.2} & \bb{48.3} & \bb{33.3} \\
$[1, 1.5, 2]$		& $[0.75, 1, 1.25, 1.5]$	& 63.6 & 46.6 & \bb{48.3} & 33.0 \\
$[1, 1.5, 2]$		& $[1, 1.25, 1.5, 1.75]$	& 63.5 & 46.4 & 48.0 & 32.8 \\
$[1, 1.75, 2.5]$	& $[0.5, 0.75, 1, 1.25]$	& 63.4 & 46.4 & 48.0 & 32.9 \\
\hline
\end{tabu}
\end{center}
\end{table}

%% file: tbl_dam.tex
\begin{table}
\begin{center}
\caption{The effects of applying the DaM convolution to Frontend module~\cite{YuKoltun2016} for the input of RGB.}\label{tab:dam}
\renewcommand{\arraystretch}{1.2} 
\begin{tabu}{c|X[c]|X[c]|X[c]|X[c]|c} 
\hline
DaM & Pixel  & Mean  & FW & Mean & Processing  \\
conv. & accu. & accu. & IoU & IoU & time ($ms$) \\
\hline\hline
None 		& 62.1 & 45.8 & 46.6 & 32.3 & 417 \\
1st 			& 63.4 & 46.7 & 48.0 & 32.9 & 467 \\
1st, 3rd, 5th 	& 63.5 & 47.0 & 48.2 & 32.9 & 470 \\
All			& \bb{63.7} & \bb{47.2} & \bb{48.3} & \bb{33.3} & 481 \\
\hline
\end{tabu}
\end{center}
\end{table}

%% file: tbl_multiscale_eval.tex
\begin{table}[t]
\begin{center}
\caption{The analysis of multi/random scale evaluation using DeepLab-based networks for the input of RGB.}\label{tab:random_multiscale_evaluation}
\renewcommand{\arraystretch}{1.2}
\begin{tabu}{c|c|c||X[c]|X[c]|X[c]|X[c]}
\hline
\multicolumn{3}{c||}{Method} & \multirow{2}{*}{Pixel}& \multirow{2}{*}{Mean} & \multirow{2}{*}{FW } & \multirow{2}{*}{Mean }  \\
\cline{1-3}
\multirow{2}{*}{DaM conv.} & \multicolumn{2}{c||}{Scaling} & \multirow{2}{*}{accu.} &  \multirow{2}{*}{accu.} & \multirow{2}{*}{IoU} & \multirow{2}{*}{IoU} \\
\cline{2-3}
& Multi & Random &  &  &  &   \\
\hline\hline
\multirow{4}{*}{-} & - & - & 63.8 & \textbf{46.2} & 48.3 & 33.7 \\
& \checkmark & - & \textbf{64.7} & 44.9 & \textbf{48.4} & \textbf{34.1}  \\
& - & \checkmark & 64.1 & 45.4 & 48.2 & 33.6  \\
& \checkmark & \checkmark & 64.6 & 45.0 & \textbf{48.4} & 33.9  \\
\cline{1-7}
\multirow{4}{*}{\checkmark} & - & - & 64.3 & \bb{47.3} & {49.0} & {34.3} \\
& \checkmark & - & \bb{65.1} & {46.7} & {49.3} & \bb{35.0}  \\
& - & \checkmark & 64.6 & 47.1 & {47.1} & {34.5}  \\
& \checkmark & \checkmark & \bb{65.1} & {47.0} & \bb{49.4} & \bb{35.0}  \\
\hline
\end{tabu}
\end{center}
\end{table}

%% file: tbl_receptive.tex
\begin{table}[t]
\begin{center}
\caption{The effects of increasing receptive fields in DeepLab.}\label{tab:dilation_analysis}
\renewcommand{\arraystretch}{1.2}
\begin{tabu}{c||X[c]|X[c]|X[c]|X[c]}
\hline
Conv. & Pixel accu. & Mean accu. & FW IoU & Mean IoU \\
\hline\hline
 - & 63.8 & 46.2 & 48.3 & 33.7 \\
$\times 2$ & 62.0 & 42.8 & 46.4 & 31.1 \\
$\times 4$ & 57.3 & 36.4 & 41.2 & 25.8 \\
DaM conv & \bb{64.3} & \bb{47.3} & \bb{49.0} & \bb{34.3} \\
\hline
\end{tabu}
\end{center}
\end{table}

%% file: fig_dataset_hand.tex
\begin{figure}[!t] \begin{center}
\begin{minipage}{0.4\linewidth}
\centerline{\includegraphics[scale=0.15]{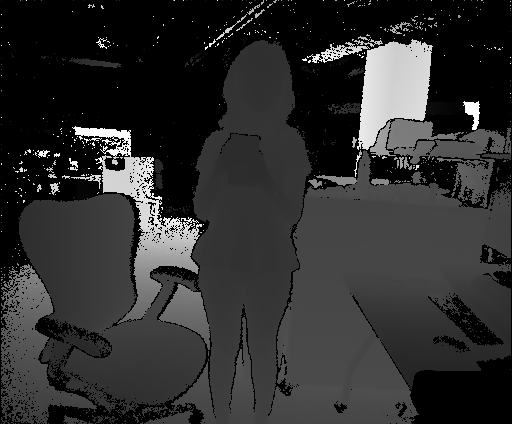}}
\centerline{\footnotesize (a)} 
\end{minipage}
\begin{minipage}{0.4\linewidth}
\centerline{\includegraphics[scale=0.15]{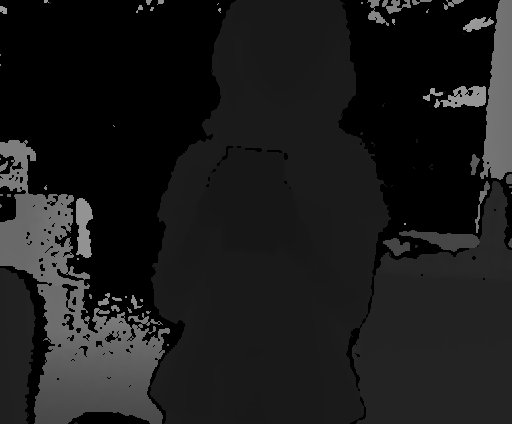}}
\centerline{\footnotesize (b)} 
\end{minipage}
\begin{minipage}{0.4\linewidth}
\centerline{\includegraphics[scale=0.15]{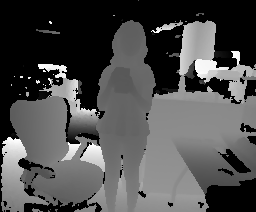}}
\centerline{\footnotesize (c)} 
\end{minipage}
   \caption{The HOI dataset. (a) A collected depth map. (b) The depth map transformed to closer distance. (c) The depth map transformed to further distance.}
\label{fig:AugDataset}
\end{center}\end{figure}

%% file: fig_dataset_dist_hand.tex
\begin{figure}[t]\begin{center}
\begin{minipage}{0.3\linewidth}
\centerline{\includegraphics[scale=0.2]{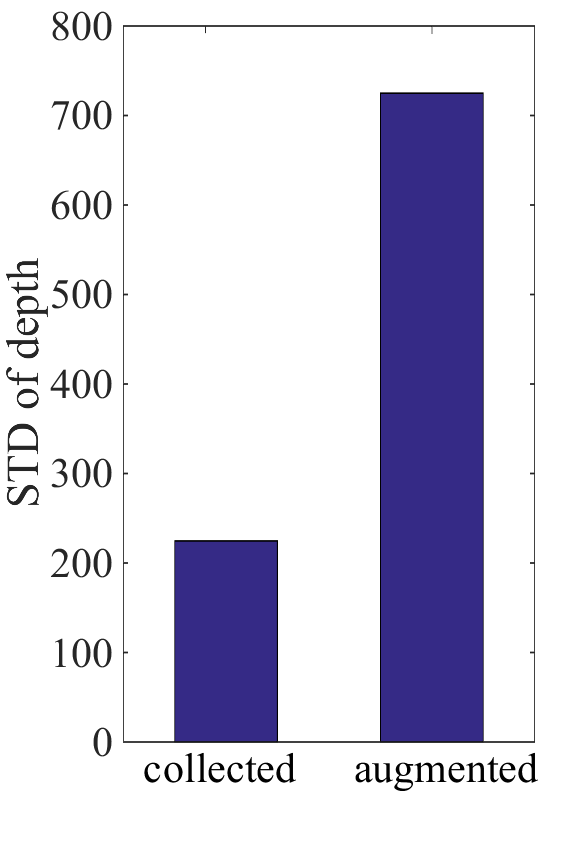}}
\centerline{\footnotesize (a)} 
\end{minipage}
\begin{minipage}{0.65\linewidth}
\centerline{\includegraphics[scale=0.2]{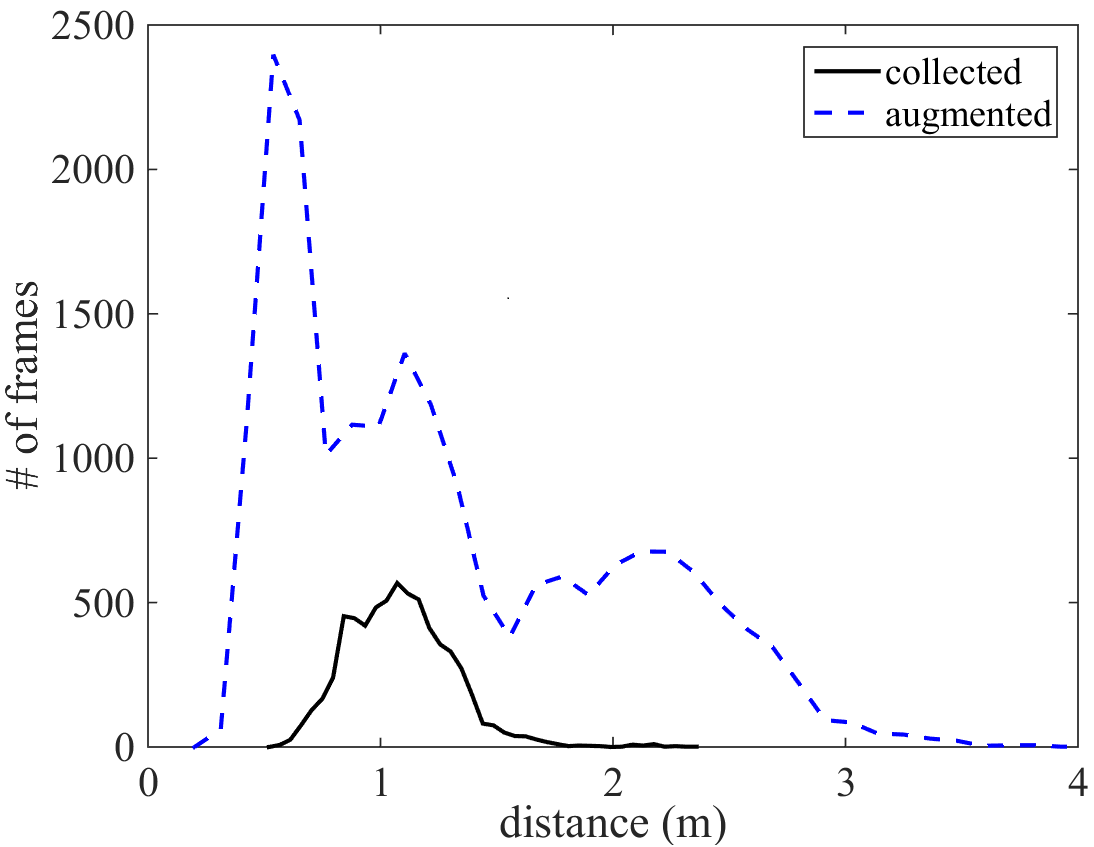}}
\centerline{\footnotesize (b)} 
\end{minipage}
\end{center}
   \caption{Analysis of the collected dataset and the augmented dataset. (a) The standard deviation of the depth of hands in $mm$.  (b) The distribution of the distance from the depth sensor to hands.}
\label{fig:distribution}
\end{figure}

%% file: tbl_result_hand.tex
\begin{table*}
\begin{center}
\caption{The quantitative results of the HOI dataset. The scores are scaled by a factor of 100. Bold face and blue color emphasize the best performance for each input and for entire cases, respectively.}
\label{tab:resultHOI}
\begin{minipage}{0.95\linewidth}
\renewcommand{\arraystretch}{1.2}
\begin{tabu}{X[c]|c||X[c]|X[c]|X[c]|X[c]|X[c]|X[c]|X[c]}
\hline
Input & Method & Precision & Recall & $F_1$ score & Pixel accu. & Mean accu. & FW IoU & Mean IoU \\
\hline\hline
\multirow{2}{*}{Depth map}  	& Frontend~\cite{YuKoltun2016} 	& 72.4 & 70.2 & 71.3 & 99.0 & 84.9 & 98.2 & 77.2 \\ 
		& Frontend + DaM conv. & \textbf{79.7} & \textbf{82.5} & \textbf{81.1} & \textbf{99.3} & \textbf{91.1} & \textbf{98.7} & \textbf{83.8} \\
\hline
\multirow{2}{*}{HHA~\cite{Gupta2014}} & Frontend~\cite{YuKoltun2016} & 76.3 & \bb{85.8} & 80.8 & 99.3 & \bb{92.7} & 98.7 & 83.5 \\
		& Frontend + DaM conv.  & \bb{83.6} & {84.1} & \bb{83.9} & \bb{99.4} & 91.9 & \bb{98.9} & \bb{85.8} \\
\hline
\end{tabu}
\end{minipage}
\end{center}
\end{table*}

%% file: fig_result_hand.tex
\begin{figure*}[!t] \begin{center}
\begin{minipage}{0.17\linewidth}
\centerline{\includegraphics[scale=0.17]{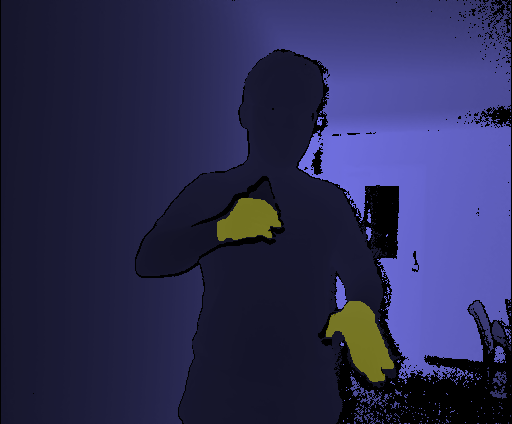}}
\end{minipage}
\begin{minipage}{0.17\linewidth}
\centerline{\includegraphics[scale=0.17]{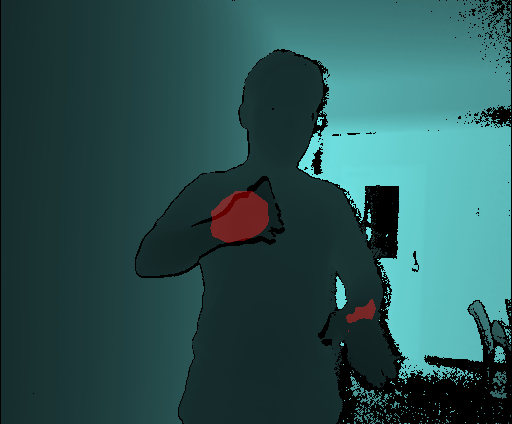}}
\end{minipage}
\begin{minipage}{0.17\linewidth}
\centerline{\includegraphics[scale=0.17]{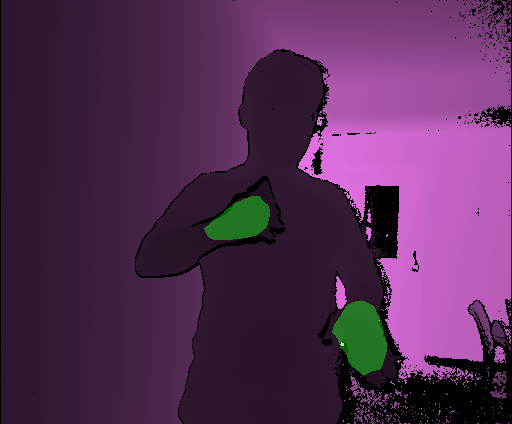}}
\end{minipage}
\begin{minipage}{0.17\linewidth}
\centerline{\includegraphics[scale=0.17]{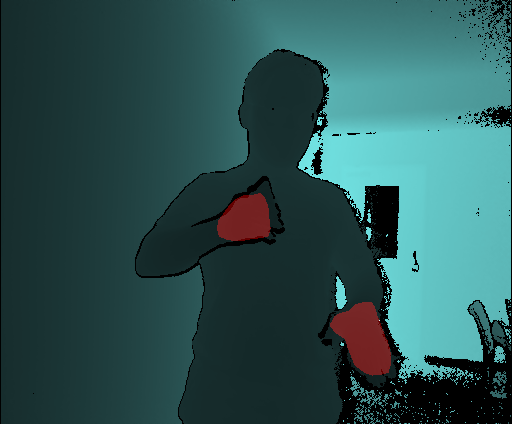}}
\end{minipage}
\begin{minipage}{0.17\linewidth}
\centerline{\includegraphics[scale=0.17]{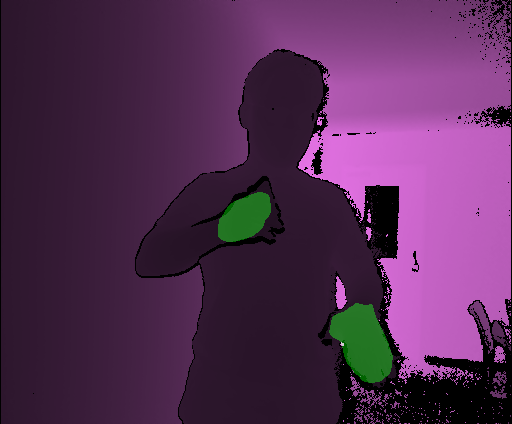}}
\end{minipage}
\\
\vspace{0.1cm}
\begin{minipage}{0.17\linewidth}
\centerline{\includegraphics[scale=0.17]{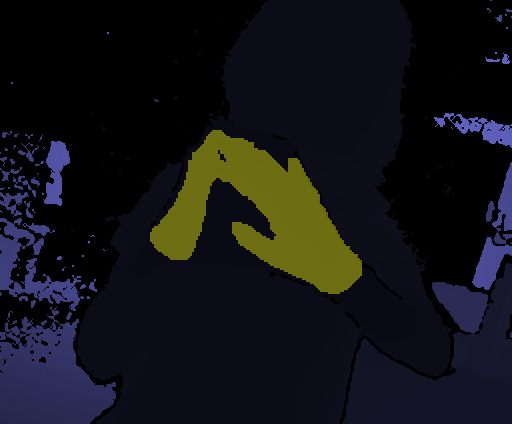}}
\end{minipage}
\begin{minipage}{0.17\linewidth}
\centerline{\includegraphics[scale=0.17]{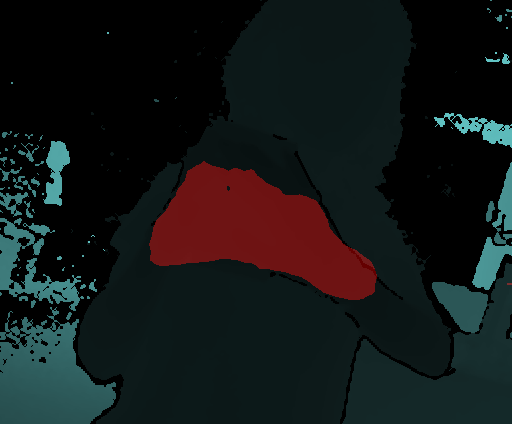}}
\end{minipage}
\begin{minipage}{0.17\linewidth}
\centerline{\includegraphics[scale=0.17]{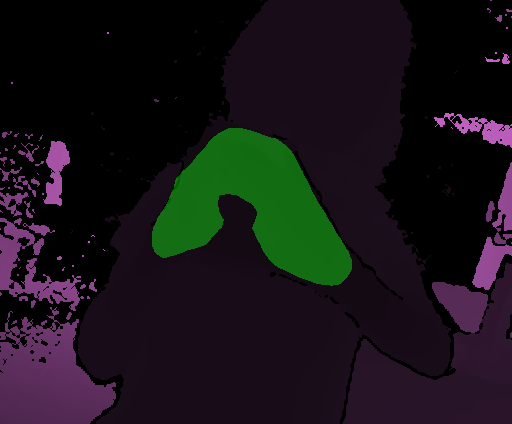}}
\end{minipage}
\begin{minipage}{0.17\linewidth}
\centerline{\includegraphics[scale=0.17]{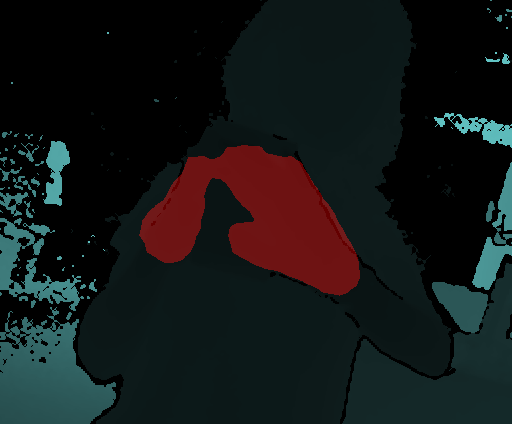}}
\end{minipage}
\begin{minipage}{0.17\linewidth}
\centerline{\includegraphics[scale=0.17]{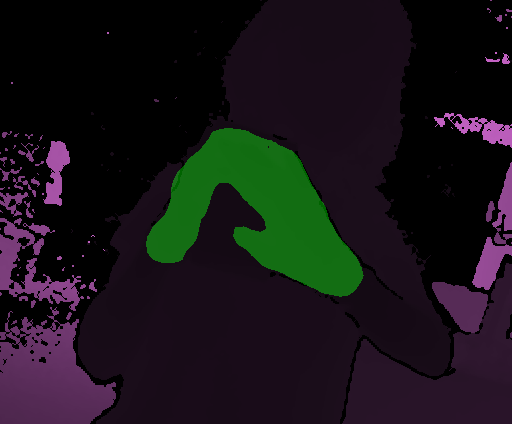}}
\end{minipage}
\\
\vspace{0.1cm}
\begin{minipage}{0.17\linewidth}
\centerline{\includegraphics[scale=0.17]{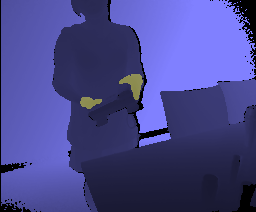}}
\centerline{\footnotesize (a): Ground truth label} 
\end{minipage}
\begin{minipage}{0.17\linewidth}
\centerline{\includegraphics[scale=0.17]{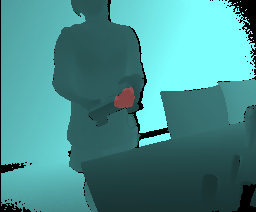}}
\centerline{\footnotesize (b): Depth} 
\end{minipage}
\begin{minipage}{0.17\linewidth}
\centerline{\includegraphics[scale=0.17]{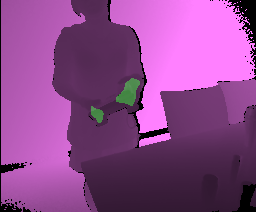}}
\centerline{\footnotesize (c): Depth with DaM conv.}
\end{minipage}
\begin{minipage}{0.17\linewidth}
\centerline{\includegraphics[scale=0.17]{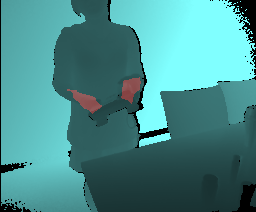}}
\centerline{\footnotesize (d): HHA} 
\end{minipage}
\begin{minipage}{0.17\linewidth}
\centerline{\includegraphics[scale=0.17]{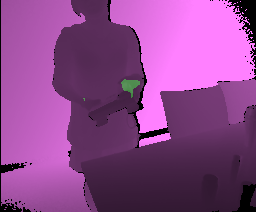}}
\centerline{\footnotesize (e): HHA with DaM conv.} 
\end{minipage}
   \caption{The qualitative comparison of the result for the HOI dataset. (a) Ground truth labels. (b) Results of Frontend module~\cite{YuKoltun2016} for the input of a depth map. (c) Results of the network with the DaM convolution for the input of a depth map. (d) Results of Frontend module~\cite{YuKoltun2016} for the input of an HHA encoded image~\cite{Gupta2014}. (e) Results of the network with the DaM convolution for the input of an HHA encoded image~\cite{Gupta2014}. The results and the ground truth labels are visualized on the depth maps with different color channels for better visualization.}
\label{fig:result}
\end{center}\end{figure*}